\documentclass[twoside,11pt]{article}
\usepackage{blindtext}

\usepackage{jmlr2e}

\usepackage{amsmath}
\usepackage{booktabs}
\usepackage{tikz}
\usetikzlibrary{arrows.meta,calc,positioning}
\usepackage{subcaption}
\usepackage{natbib}
\usepackage{mathrsfs}
\usetikzlibrary{arrows.meta,positioning,fit}
\usetikzlibrary{arrows.meta,positioning,calc,fit}

\newcommand{\Com}{\mathcal C}
\newcommand{\aaa}{a}
\newcommand{\bbb}{b}
\newcommand{\adj}{\mathfrak{N}}
\newcommand{\ZZ}{Z}

\newtheorem{assumption}{Assumption}

\usepackage{lastpage}
\begin{document}

\title{A Measure-Geometric Theory of Robustness in Machine Learning}

\author{\name Cheng-Chao Huang \email chengchao@njis.ac.cn \\
       \addr Nanjing Institute of Software Technology\\
       Nanjing 211135, China\\
       }

\editor{}

\maketitle

\begin{abstract}%
This paper develops a foundational measure-geometric theory of robustness in machine learning. Robustness is formulated as escape from a retained task state relative to a reference measure on a data-bearing manifold, a perturbation gauge, and an admissible realization mechanism. 
The theory distinguishes the structural possibility of escape on the reference manifold from its ambient realization under mechanism-dependent observable readouts, while placing both within a unified geometric account. It introduces a mesoscopic organization of escape geometry, through which vulnerable mass, finite-scale morphology, delivered progress, and branch coherence are characterized by explicit laws as linked but non-identical robustness phenomena. By recasting robustness as the geometric formation, organization, realization, and persistence of escape, the theory provides a structural basis for analysis beyond scalar criteria.
\end{abstract}

\begin{keywords}
machine learning robustness, measure-geometric theory, formal robustness analysis, adversarial robustness, data manifold geometry
\end{keywords}

% \section{Introduction}
% \blindmathpaper
% Here is a citation \citep{chow:68}.

\section{Introduction}
\label{sec:introduction}

Robustness is a basic requirement for machine learning systems that must retain a task-relevant state
under admissible changes.
It expresses whether learned behavior remains consistent when the input, environment,
representation, or mode of intervention varies within a prescribed scope.
This issue is central to modern machine learning, since deployed models are expected to operate not
only on ideal samples, but also under natural variation, distributional shift, semantic editing, sensor
noise, model uncertainty, and adversarial probing, with adversarial examples and robust optimization
providing one of the most developed formal settings~\citep{szegedy2014intriguing,goodfellow2015explaining,madry2018towards}.

In practice, this requirement is often summarized by quantities such as margins, distances to retention
boundaries, robust accuracies, certificates, or attack success rates.
These quantities are indispensable because they make robustness measurable, trainable, and testable.
Yet each of them records only one projection of a richer underlying structure.
They do not, by themselves, reveal how vulnerable regions and their associated data mass are
generated as the perturbation budget increases.
They also do not reveal how a concrete attack procedure uses local model information to produce a
realized adversarial trajectory that reaches failure.
A substantial theoretical literature has gone beyond operational summaries by relating adversarial
robustness to the geometry of the data distribution, the local geometry of decision boundaries,
universal adversarial perturbations, and robust or non-robust features
~\citep{fawzi2018analysis,moosavi2018robustness_universal,stutz2019disentangling,ilyas2019features}.
These views identify vulnerability as a structural relation among the data distribution, perturbation
geometry, decision geometry, and learned features, while leaving comparatively implicit the intermediate
organization from structural sources to the formation and realization of vulnerability.

Within this broad geometric orientation, we develop a foundational measure-geometric theory of
robustness as escape.
The term geometric reflects that loss of robustness is organized by the boundary geometry of the
retained state, the local openings through which vulnerable regions are generated, and the ambient
paths through which escape is realized.
The term measure reflects that these objects enter the theory through their interaction with the
reference measure whose mass is under consideration.
In this theory, escape encompasses boundary crossing, the generation of vulnerable regions on the reference manifold, their measurement under the reference measure, and their realization through admissible ambient paths.
The perturbation gauge specifies how admissible changes are measured and thereby induces the
escape-energy field on the reference manifold.
The admissible realization mechanism specifies the ambient paths and observable readouts through which
escape is carried out.
The theory therefore organizes escape geometry at a mesoscopic level, treating vulnerable mass,
finite-scale morphology, delivered progress, and branch coherence as linked but non-identical aspects
of robustness.
Thus robustness is a measure-geometric relation among the retained task state, the data-bearing
geometry, the perturbation gauge, and the admissible realization mechanism.

This perspective separates two questions that are often conflated.
The first concerns escape structure on the data-bearing reference geometry.
Where is escape present at a given energy scale, and how does vulnerable mass change as the energy
threshold varies?
The second concerns ambient realization.
Given that such escape is geometrically present, can a concrete mechanism access and deliver it through
the observable readouts available along its realized path?
The former concerns the formation and organization of vulnerability. The latter concerns its realization through a mechanism-dependent geometry. This separation echoes a fundamental distinction in theoretical computer science between the structure of a solution space and the ability of a given procedure to obtain a witness.
A model may contain substantial escape structure at the relevant budget scale that a particular
realization mechanism does not efficiently access.
Conversely, a realized failure may reflect not merely that the boundary is close, but that the realized
trajectory is well aligned with the ambient geometry that measures remaining escape.

\emph{Contributions and scope.}
This work formulates a foundational measure-geometric theory of robustness as escape.
The theory develops a structural calculus that separates escape formation on a measured reference geometry from escape realization by admissible ambient mechanisms.
It turns minimal perturbation into an escape-energy field, derives componentwise laws for vulnerable-mass growth and finite-scale occupancy, and uses ambient extensions and observable readouts to describe mechanism-accessible delivery of escape.
Its scope is structural rather than algorithmic.
It provides a common language for geometric, measure-theoretic, verification-oriented, and attack-oriented analyses, while allowing concrete learning problems to refine the same escape objects with additional regularity, computability, and statistical structure.

\emph{Organization.}
Section~\ref{sec:mathematical_preliminaries} fixes the measure-theoretic, geometric, and nonsmooth
analytic conventions used throughout the paper.
Section~\ref{sec:reference_escape_structure} introduces the reference escape structure and the
escape-energy field on the data-bearing geometry.
Section~\ref{sec:shell_energy_axis} analyzes how escape geometry is projected into energy-axis growth
and finite-scale occupancy.
Section~\ref{sec:realized_escape_dynamics} develops the ambient dynamics by which escape is realized
through concrete mechanisms.
Section~\ref{sec:ml_robustness_geometry} interprets the theory in machine-learning robustness terms and relates it to existing geometric views.
Section~\ref{sec:concluding_remarks} concludes.

\section{Mathematical Preliminaries}
\label{sec:mathematical_preliminaries}

Measure-theoretic, rectifiable, integral, distributional-derivative, and nonsmooth first-order conventions are
fixed in this section.
All sets and maps are assumed measurable whenever they are used as arguments of measures,
pushforwards, or integrals.
Equalities and inclusions explicitly marked by a measure are understood modulo the corresponding null
sets. 

\subsection{Notation, Measures, and Null Sets}
\label{subsec:prelim_basic_notation}

For $\alpha\in\mathbb R$, write $\alpha_+:=\max\{\alpha,0\}$.
For a set $A$, write $\mathbf 1_A$ for its indicator function.
For $k\ge1$, write $\mathcal L^k$ for $k$-dimensional Lebesgue measure on $\mathbb R^k$.
For measurable $E\subset\mathbb R$, write $|E|:=\mathcal L^1(E)$.

For Euclidean space $\mathbb R^q$, write
\[
\mathbb B_r^q:=\{z\in\mathbb R^q:\|z\|<r\},
\qquad
\mathbb S^{q-1}:=\{z\in\mathbb R^q:\|z\|=1\}.
\]
Thus $\mathbb B_r^q$ is centered at the origin.
For $a\in\mathbb R^q$, write $\mathbb B_r^q(a):=a+\mathbb B_r^q$.

If $I$ and $J$ are intervals, write $I\Subset J$ when the closure of $I$ is compact and contained in
$J$.
For nonnegative quantities $M$ and $N$ depending on an asymptotic parameter, write $M\asymp N$ if there
exist constants $0<c\le C<\infty$, independent of that parameter, such that $cN\le M\le CN$.
The constants may depend on fixed structural data.

Let $(X,d)$ be a metric space equipped with a Borel measure $\nu$.
For nonempty subsets $A,B\subset X$, write
$d(A,B):=\inf\{d(a,b):a\in A,\ b\in B\}$.
For a measurable set $A\subset X$, write $\nu\!\lfloor_A$ for the restriction of $\nu$ to $A$.
If $\Phi:X\to Y$ is measurable, write $\Phi_\#\nu$ for the pushforward measure, defined by
$\Phi_\#\nu(E):=\nu(\Phi^{-1}(E))$.

For two measures $\nu$ and $\rho$ on the same measurable space, write $\nu\ll\rho$ if $\nu$ is
absolutely continuous with respect to $\rho$.
When $\nu\ll\rho$, the Radon--Nikodym density is denoted by $d\nu/d\rho$.

For measurable sets $A,B\subset X$, write $A\subseteq_\nu B$ if $\nu(A\setminus B)=0$.
Write $A=_\nu B$ if both $A\subseteq_\nu B$ and $B\subseteq_\nu A$.
Thus containment and equality are understood up to $\nu$-null sets.

A measurable set $A\subset X$ is called $\nu$-essentially connected if every measurable partition
$A=A_1\cup A_2$ with $A_1\cap A_2=\varnothing$ and $\nu(A_1),\nu(A_2)>0$ satisfies $d(A_1,A_2)=0$.
Let $\mathcal X$ be a class of measurable subsets of $X$.
A set $A\in\mathcal X$ is called $\subseteq_\nu$-maximal in $\mathcal X$ if, whenever
$B\in\mathcal X$ and $A\subseteq_\nu B$, one also has $B\subseteq_\nu A$.

Whenever a measurable finite- or countable-label map is used to define cells, a modulo-null
decomposition of each positive-measure label set into at most countably many $\nu$-essentially
connected $\subseteq_\nu$-maximal measurable pieces is fixed whenever such a decomposition is
invoked.

\subsection{Rectifiable Geometry and Coordinate Charts}
\label{subsec:prelim_measure_geometry}

For $k\ge0$, write $\mathcal H^k$ for $k$-dimensional Hausdorff measure
~\citep[Sec.~2.10]{federer_gmt}.
If $\mathcal N\subset\mathbb R^n$ is an embedded $m$-dimensional $C^1$ submanifold, then
$\mathcal H^m\!\lfloor_{\mathcal N}$ is the intrinsic volume measure on $\mathcal N$, and
$T_x\mathcal N$ denotes the tangent space at $x\in\mathcal N$.

A set $E\subset\mathbb R^n$ is countably $k$-rectifiable if there exist countably many Lipschitz maps
$f_i:E_i\subset\mathbb R^k\to\mathbb R^n$ such that
\[
\mathcal H^k\!\left(E\setminus\bigcup_{i=1}^\infty f_i(E_i)\right)=0 .
\]
Equivalently, up to an $\mathcal H^k$-null set, $E$ is covered by countably many Lipschitz
$k$-dimensional images~\citep[Secs.~3.2.14--3.2.15]{federer_gmt}.

A set contained, up to an $\mathcal H^{n-1}$-null set, in countably many locally Lipschitz hypersurface
pieces admits approximate tangent hyperplanes at $\mathcal H^{n-1}$-a.e.\ point
~\citep[Secs.~3.2.16--3.2.19]{federer_gmt}.
Whenever a normal field appears, a measurable choice of unit normal is fixed on the corresponding
a.e.\ tangent hyperplane field.

If $E$ is countably $k$-rectifiable and a finite Borel measure $\nu$ satisfies
$\nu\ll\mathcal H^k\!\lfloor_E$, then
$d\nu/d(\mathcal H^k\!\lfloor_E)$ denotes its density on $E$.
Uniform density bounds mean that this Radon--Nikodym derivative is bounded above and below by positive
constants on the set under consideration.

A map $\chi:U\to Z$ between metric spaces is bi-Lipschitz if it is bijective onto its image and both
$\chi$ and $\chi^{-1}$ are Lipschitz.
When a finite Borel measure $\nu$ is represented in a bi-Lipschitz chart
$\chi:U\subset\mathbb R^m\to\mathbb R^n$, a coordinate density $w$ means that, for every measurable
$A\subset U$,
\[
\nu(\chi(A))=\int_A w(u)\,d\mathcal L^m(u).
\]
In product coordinates $U=S\times W\subset\mathbb R^{m-q}\times\mathbb R^q$, the coordinate product
measure is $\mathcal L^{m-q}\otimes\mathcal L^q$.
Uniform upper and lower bounds on $w$ convert coordinate-volume estimates into measure estimates up to
fixed multiplicative constants.

\subsection{Integral and Distributional-Derivative Conventions}
\label{subsec:prelim_integral_distributional}

Tonelli's theorem~\citep[Thm.~1.7.18]{tao2011measure} is used in the following form.
If $H:X\times Y\to[0,\infty]$ is measurable on a product measure space, then
\[
\int_{X\times Y}H\,d(\nu\otimes\rho)
=
\int_X\int_Y H(x,y)\,d\rho(y)\,d\nu(x)
=
\int_Y\int_X H(x,y)\,d\nu(x)\,d\rho(y).
\]
The dominated convergence theorem~\citep[Thm.~1.4.49]{tao2011measure} is used in its standard form:
if $H_j\to H$ pointwise a.e.\ with respect to $\nu$ and $|H_j|\le G$ for some $G\in L^1(\nu)$, then
$\int H_j\,d\nu\to\int H\,d\nu$.

The coarea formula~\citep[Thm.~3.2.22]{federer_gmt} is used for Lipschitz maps with a.e.\ defined
Jacobians.
If $\Psi:\mathbb R^m\to\mathbb R^k$ is Lipschitz with $m\ge k$ and $g\ge0$ is measurable, then
\[
\int_{\mathbb R^m} g(x)\,J_k\Psi(x)\,dx
=
\int_{\mathbb R^k}\int_{\Psi^{-1}(y)} g(x)\,d\mathcal H^{m-k}(x)\,dy .
\]
Here $J_k\Psi$ denotes the $k$-dimensional Jacobian.

The area formula~\citep[Thm.~3.2.3]{federer_gmt} is used for Lipschitz images.
For a Lipschitz map $\Phi:E\subset\mathbb R^m\to\mathbb R^n$ with $m\le n$, define the multiplicity
function by $N(\Phi,E,y):=\#\{x\in E:\Phi(x)=y\}$.
For nonnegative measurable $g$,
\[
\int_E g(x)\,J_m\Phi(x)\,dx
=
\int_{\mathbb R^n}
\sum_{x\in E\cap\Phi^{-1}(y)} g(x)\,d\mathcal H^m(y),
\]
whenever the relevant Jacobian and integrability conditions are satisfied.
A uniform bound on $N(\Phi,E,\cdot)$ controls the overlap of the parametrization.

For an open interval $I\subset\mathbb R$, write
$\mathcal D(I):=C_c^\infty(I)$ and $\mathcal D'(I):=\mathcal D(I)^\ast$.
Also write $L^1_{\mathrm{loc}}(I)$ for the functions integrable on every compact subinterval of $I$.
If $g,h\in L^1_{\mathrm{loc}}(I)$, the inequality $g'(t)\le h(t)$ in $\mathcal D'(I)$ means that
\[
-\int_I g(t)\varphi'(t)\,dt
\le
\int_I h(t)\varphi(t)\,dt
\]
for every nonnegative $\varphi\in C_c^\infty(I)$.
Similarly, $-g'(t)\le h(t)$ in $\mathcal D'(I)$ means
\[
\int_I g(t)\varphi'(t)\,dt
\le
\int_I h(t)\varphi(t)\,dt
\]
for every nonnegative $\varphi\in C_c^\infty(I)$.

If a bounded-variation representative of $g$ satisfies
$g(a)-g(b)\le\int_a^b h(t)\,dt$ for every $a<b$ in $I$, this cumulative estimate is written as
$-g'(t)\le h(t)$ in $\mathcal D'(I)$.

\subsection{Clarke First-Order Geometry}
\label{subsec:prelim_clarke_geometry}

The Clarke derivative and subdifferential are used for locally Lipschitz scalar fields
~\citep[Sec.~2.1]{clarke_optimization}.
For nonzero $u,v\in\mathbb R^n$, define
\[
d_{\angle}(u,v)
:=
\arccos\left(\frac{u\cdot v}{\|u\|\,\|v\|}\right)\in[0,\pi].
\]

Let $f$ be locally Lipschitz on an open set $U\subset\mathbb R^n$.
The Clarke generalized directional derivative of $f$ at $z\in U$ in direction $v\in\mathbb R^n$ is
\[
f^\circ(z;v)
:=
\limsup_{\substack{y\to z\\ h\downarrow0}}
\frac{f(y+hv)-f(y)}{h}.
\]
The Clarke subdifferential is
\[
\partial^{\mathrm C}f(z)
:=
\{\zeta\in\mathbb R^n:\ f^\circ(z;v)\ge \zeta\cdot v
\ \text{for all }v\in\mathbb R^n\}.
\]
For locally Lipschitz $f$, the set $\partial^{\mathrm C}f(z)$ is nonempty, compact, and convex.
Moreover, the standard support-function representation of the Clarke subdifferential
~\citep[Prop.~2.1.2]{clarke_optimization} gives
\[
f^\circ(z;v)=\max_{\zeta\in\partial^{\mathrm C}f(z)} \zeta\cdot v .
\]

When a locally Lipschitz function is defined on a $C^1$ embedded submanifold, its Clarke directional
derivative in tangent directions is understood in local $C^1$ charts.
Equivalently, if $\chi:O\subset\mathbb R^m\to\mathcal N$ is a local chart, $x=\chi(u)$, and
$v=D\chi(u)\eta\in T_x\mathcal N$, then the notation $f^\circ(x;v)$ means
\[
(f\circ\chi)^\circ(u;\eta).
\]
All tangent-space Clarke objects below are taken in this intrinsic chartwise sense.

The Clarke steepness of $f$ at $z$ is
\[
s_f(z):=\operatorname{dist}(0,\partial^{\mathrm C}f(z)).
\]
On the noncritical set $\{z:s_f(z)>0\}$, define
\[
w_f(z):=
-\frac{\operatorname{proj}_{\partial^{\mathrm C}f(z)}(0)}{s_f(z)}
\in\mathbb S^{n-1}.
\]
Here $\operatorname{proj}_{\partial^{\mathrm C}f(z)}(0)$ is the Euclidean projection of the origin onto
the compact convex set $\partial^{\mathrm C}f(z)$.
Points with $s_f(z)=0$ are called Clarke-critical, and $w_f$ is used only on $\{s_f>0\}$.

\begin{definition}[Clarke transverse spread]
\label{def:transverse_clarke_spread}
Let $f$ be locally Lipschitz on an open set $U\subset\mathbb R^n$, and let $z\in U$ satisfy
$s_f(z)>0$.
Write $\zeta_f^\ast(z):=\operatorname{proj}_{\partial^{\mathrm C}f(z)}(0)$.
The Clarke transverse spread of $f$ at $z$ is
\[
\operatorname{spr}_f(z)
:=
\sup_{\zeta\in\partial^{\mathrm C}f(z)}
\left\|
\Pi_{\zeta_f^\ast(z)^\perp}\zeta
\right\|,
\]
where $\Pi_{\zeta_f^\ast(z)^\perp}$ denotes the orthogonal projection onto the hyperplane perpendicular
to $\zeta_f^\ast(z)$.
The spread is said to vanish at $z$ if $\operatorname{spr}_f(z)=0$.
\end{definition}

Vanishing Clarke transverse spread means that every element of $\partial^{\mathrm C}f(z)$ has zero component
orthogonal to the minimal-norm element $\zeta_f^\ast(z)$.
It is automatic when $\partial^{\mathrm C}f(z)$ is a singleton.

\begin{lemma}[Cone descent under vanishing Clarke transverse spread]
\label{lem:cone_descent_vanishing_clarke_spread}
Let $f$ be locally Lipschitz on an open set $U\subset\mathbb R^n$, and let $z\in U$ satisfy
$s_f(z)>0$.
Assume that $f$ has vanishing Clarke transverse spread at $z$.
Then, for every nonzero $u\in\mathbb R^n$ with
\[
d_{\angle}(u,w_f(z))\le \frac{\pi}{2},
\]
one has
\[
-f^\circ(z;u)
\ge
s_f(z)\|u\|
\cos d_{\angle}(u,w_f(z)).
\]
\end{lemma}

\begin{proof}
Set $C:=\partial^{\mathrm C}f(z)$, $\zeta^\ast:=\operatorname{proj}_{C}(0)$, and
$s:=\|\zeta^\ast\|=s_f(z)>0$.
Then $w_f(z)=-\zeta^\ast/s$.
The projection property, namely the variational characterization of the Euclidean projection onto a closed convex set, gives
$\zeta\cdot\zeta^\ast \ge s^2$ for every $\zeta\in C$.
Since the Clarke transverse spread vanishes, each $\zeta\in C$ is parallel to $\zeta^\ast$, hence
$\zeta=a\zeta^\ast$ with $a\ge1$.

Let $e\in\mathbb S^{n-1}$ and $\alpha:=d_{\angle}(e,w_f(z))\le\pi/2$.
Then $\zeta^\ast\cdot e=-s\cos\alpha\le0$, so for every $\zeta=a\zeta^\ast\in C$,
$\zeta\cdot e\le\zeta^\ast\cdot e=-s\cos\alpha$.
By the support-function formula,
\[
f^\circ(z;e)=\max_{\zeta\in C}\zeta\cdot e\le -s\cos\alpha .
\]
Thus $-f^\circ(z;e)\ge s\cos\alpha$.
The claim for nonzero $u$ follows by applying this to $e=u/\|u\|$ and using positive homogeneity.
\end{proof}

For an absolutely continuous path $\xi:[a,b]\to U$, the composition
$t\mapsto f(\xi(t))$ is absolutely continuous. Moreover, the standard
pathwise Clarke estimate gives
\[
\frac{d}{dt}f(\xi(t))
\le
f^\circ(\xi(t);\dot\xi(t))
\qquad
\text{for a.e. }t\in[a,b].
\]
Indeed, at a.e.\ $t$ one has
$\xi(t+h)=\xi(t)+h\dot\xi(t)+o(h)$, and the local Lipschitz continuity of
$f$ reduces the derivative of $f\circ\xi$ to the defining upper limit in
$f^\circ$.

%%%%%%%%%%%%%%%%%%%%%%%%%%%%%%%%%%%%%%%
%%%%%%%%%%%%%%%%%%%%%%%%%%%%%%%%%%%%%%%

\section{Reference Escape Structure}
\label{sec:reference_escape_structure}

Robustness is cast here as escape.
A point on the reference manifold is vulnerable to the extent that an admissible perturbation can carry it to the retention boundary under the chosen gauge.
The resulting minimal escape energy turns vulnerability into a scalar geometry on $\mathcal M$.
The geometric structures generated by this field furnish the primitive objects of the analysis.

\subsection{Reference Setting and Escape-Energy Field}
\label{subsec:reference_setting}

Consider a problem defined on an ambient Euclidean space $\mathbb{R}^n$.
The system of interest is specified by a locally Lipschitz real-valued function
$f:\mathbb{R}^n\to\mathbb{R}$.
The associated superlevel region is
$\Omega:=\{x\in\mathbb{R}^n:f(x)>0\}$,
and we write
$\partial\Omega:=\{x\in\mathbb{R}^n:f(x)=0\}$
for the zero-level decision boundary, assumed throughout to coincide with the topological boundary of $\Omega$.
Then $\Omega$ is open, and $\partial\Omega$ is closed, hence also Borel.

Assume the system is observed through a compact $m$-dimensional $C^1$ embedded submanifold
$\mathcal M\subset\mathbb{R}^n$,
which serves as a low-dimensional reference structure.
The manifold $\mathcal M$ is endowed with a Borel probability measure $\mu$ that is absolutely
continuous with respect to the $m$-dimensional Hausdorff measure $\mathcal H^m$ on $\mathcal M$.

The reference manifold $\mathcal M$ is the geometric carrier on which escape is defined and measured.
At the formal level, it provides tangent spaces and the intrinsic volume
$\mathcal H^m\!\lfloor_{\mathcal M}$ on which the subsequent measurable constructions are posed.
At the geometric level, it supplies the background against which the escape-energy field may develop
nontrivial local degeneracy or higher-order contact with the reference geometry.

We introduce a variational specification of perturbation-induced escape over admissible perturbation paths.
For each $x\in\mathcal M$, let $\mathcal A_x$ be a family of admissible paths
$\xi:[0,\infty)\to\mathbb R^n$ that are absolutely continuous on every compact interval,
satisfy $\xi(0)=x$, and reach the boundary $\partial\Omega$ in finite time.
More precisely, for $\xi\in\mathcal A_x$, define its first hitting time of $\partial\Omega$ by
\[
\tau_{\partial\Omega}(\xi):=\inf\{t\ge 0:\xi(t)\in\partial\Omega\},
\]
and require $\tau_{\partial\Omega}(\xi)<\infty$.
The family $\mathcal A_x$ specifies the admissible perturbation mechanism.

To measure perturbation size, fix a measurable function
\[
p:\mathbb R^n\times\mathbb R^n\to[0,\infty)
\]
such that $p(z,z)=0$ and
$p(z_0,z_2)\le p(z_0,z_1)+p(z_1,z_2)$ for all $z,z_0,z_1,z_2\in\mathbb R^n$.
The value $p(z_1,z_2)$ is interpreted as the perturbation gauge from $z_1$ to $z_2$.
The vanishing condition fixes the zero gauge at a point, while the subadditivity is the basic concatenation law for perturbation transport.
In particular, for an initial point $x\in\mathcal M$ and an admissible path $\xi\in\mathcal A_x$, the quantity
$p\bigl(x,\xi(\tau_{\partial\Omega}(\xi))\bigr)$
measures the total perturbation energy required to reach the boundary along that realized escape.
This induces a scalar escape energy at each $x\in\mathcal M$, given by the minimal perturbation required to reach $\partial\Omega$ along admissible paths.

\begin{definition}[Escape-energy field]
\label{def:escape_cost_variational}
The escape-energy field is the real-valued function
\[
\phi(x)
:=
\inf\bigl\{
p\bigl(x,\xi(\tau_{\partial\Omega}(\xi))\bigr):\ \xi\in\mathcal A_x
\bigr\},
\qquad x\in\mathcal M.
\]
\end{definition}

The following basic analytic assumptions ensure that the geometric objects induced by $\phi$ are well defined
$\mu$-a.e.\ and admit measurable representatives.
Unless explicitly stated otherwise, every assumption introduced below is standing onward.

\begin{assumption}[Basic regularity]
\label{ass:basic_regularity}
The variational specification is proper on $\mathcal M$, in the sense that
$\mathcal A_x\neq\varnothing$ and $\phi(x)<\infty$ for all $x\in\mathcal M$, and the escape-energy
field $\phi$ is locally Lipschitz on $\mathcal M$.
\end{assumption}

For $\delta>0$, a point $x\in\mathcal M$ is called $\delta$-escapable if $\phi(x)\le \delta$,
namely if it admits an admissible path to $\partial\Omega$ with endpoint perturbation at most $\delta$.
The scalar $\delta$ therefore acts as an energy threshold, and $\phi$ induces the corresponding energy
scale through its sublevel sets.

\begin{definition}[Escape set]
\label{def:escape_set}
For $\delta>0$, the escape set at energy $\delta$ is
\[
X_\delta
\;:=\;
\{x\in\mathcal M:\ \phi(x)\le \delta\}.
\]
\end{definition}

The escape set $X_\delta$ records the portion of the reference manifold $\mathcal M$ that has become
accessible to escape at energy level $\delta$.
It is therefore a manifold-level notion of vulnerability, which identifies those data-bearing points on
$\mathcal M$ from which the decision boundary can be reached within the prescribed energy threshold,
rather than an ambient subset of $\mathbb R^n$ viewed independently of the data geometry.
Accordingly, $\mu(X_\delta)$ is the vulnerable mass revealed at energy $\delta$, measured with respect
to the reference distribution on $\mathcal M$.

\subsection{Transverse Rise and Escape Components}
\label{subsec:transverse_rise_escape_components}

To identify the local geometric units that govern escape, we first extract from the escape-energy field
its intrinsic energy-neutral tangential structure.
For $x\in\mathcal M$, define the energy-neutral lineality subspace by
\[
\mathrm{Lin}(x)
:=
\{v\in T_x\mathcal M:\phi^\circ(x;v)=0\ \text{and}\ \phi^\circ(x;-v)=0\}.
\]
This is the tangential subspace along which the escape-energy field is locally neutral to first order in both directions.
Since $x\mapsto\mathrm{Lin}(x)$ admits a Borel representative as a Grassmannian-valued map, the flat dimension $d_{\mathrm{fl}}(x):=\dim \mathrm{Lin}(x)$ is measurable~\citep[Thm.~14.8]{rockafellar_wets_va}. 
The flat dimension records the size of the first-order neutral tangential geometry, while the local
escape geometry also depends on the transverse rise of $\phi$ away from this subspace.

For $d\in\{0,1,\dots,m-1\}$, a transverse opening sector in $\mathbb R^{m-d}$ is a measurable unit-truncated cone
$\mathbb E^{m-d}\subset \mathbb B_1^{m-d}$ such that
$0\in\mathbb E^{m-d}$ and 
$0<\mathcal L^{m-d}(\mathbb E^{m-d})<\infty$.
Here unit-truncated cone means that $a z\in\mathbb E^{m-d}$ whenever $z\in\mathbb E^{m-d}$, $a\ge0$, and
$a z\in\mathbb B_1^{m-d}$.
For $r>0$, write
\[
\mathbb E_r^{m-d}:=r\mathbb E^{m-d}.
\]
The set $\mathbb E_r^{m-d}$ records the transverse opening through which the shell base generates
higher-energy points.

\begin{definition}[$d$-adapted rise chart at energy $\delta$]
\label{def:d_adapted_rise_chart}
Fix $\delta>0$ and $d\in\{0,1,\dots,m-1\}$.
An adapted rise chart at energy $\delta$ is a bi-Lipschitz map
\[
\chi:S\times \mathbb E_r^{m-d}\to\mathcal M,
\]
where $S\subset\mathbb R^d$ is measurable with $0<\mathcal L^d(S)<\infty$, and
$\mathbb E^{m-d}$ is a transverse sector, such that
\[
\chi(S\times\{0\})
\subset
\{x\in\mathcal M:\ \phi(x)=\delta,\ d_{\mathrm{fl}}(x)=d\}.
\]
\end{definition}
Thus the subset $S\times\{0\}$ parametrizes a $d$-dimensional portion of
$\{x\in\mathcal M:\ \phi(x)=\delta,\ d_{\mathrm{fl}}(x)=d\}$,
while the factor $\mathbb E_r^{m-d}$ records the transverse opening through which the thin shell is generated.
We say that an adapted rise chart passes through $x$ if $x\in\chi(S\times\{0\})$.

\begin{assumption}[Local transverse rise structure]
\label{ass:local_transverse_rise_escape_component}
For $\mathcal L^1$-a.e.\ $\delta>0$ and each $d\in\{0,1,\dots,m-1\}$, for $\mathcal H^d$-a.e.\
$x\in\{x\in\mathcal M:\ \phi(x)=\delta,\ d_{\mathrm{fl}}(x)=d\}$,
there exists an adapted rise chart
\[
\chi:S\times \mathbb E_r^{m-d}\to\mathcal M
\]
through $x$, and there exist an exponent $\beta\ge1$, constants
$0<\underline c\le \overline c<\infty$ and $0<\underline w\le \overline w<\infty$, 
and a coordinate density
$w:S\times \mathbb E_r^{m-d}\to[0,\infty)$ of $\mu$ in the chart $\chi$ such that:
\begin{enumerate}
\item[\textnormal{(i)}] \textnormal{Rise bound.}
After modification on a $(\mathcal L^d\otimes\mathcal L^{m-d})$-null set,
\[
\underline c\,\|z\|^\beta
\le
\phi(\chi(y,z))-\phi(\chi(y,0))
\le
\overline c\,\|z\|^\beta
\]
for $(\mathcal L^d\otimes\mathcal L^{m-d})$-a.e.\ $(y,z)\in S\times \mathbb E_r^{m-d}$.

\item[\textnormal{(ii)}] \textnormal{Nondegenerate density.}
For every measurable $A\subset S\times \mathbb E_r^{m-d}$,
\[
\mu(\chi(A))
=
\int_A w(y,z)\,d(\mathcal L^d\otimes\mathcal L^{m-d})(y,z),
\]
with
$\underline w\le w(y,z)\le \overline w$
for $(\mathcal L^d\otimes\mathcal L^{m-d})$-a.e.\ $(y,z)\in S\times \mathbb E_r^{m-d}$.
\end{enumerate}
\end{assumption}

Assumption~\ref{ass:local_transverse_rise_escape_component} provides a local product model for the
escape geometry near $\{\phi=\delta\}$.
In the adapted chart, the shell base is represented by the slice $z=0$, and the escape-energy field
rises through the transverse opening coordinate $z$.
Condition~(i) captures the local law by which $\phi$ increases away from the shell base, with the
exponent $\beta$ recording the order of that increase.
Condition~(ii) rules out measure-theoretic degeneracy in these coordinates.
Thus the local escape geometry is governed by the transverse growth of $\phi$ along the generated opening.
The case $d_{\mathrm{fl}}=m$ means that there is no intrinsic transverse direction left inside
$\mathcal M$.
Such points represent plateau-critical behavior rather than regular shell generation, and are excluded
from the transverse-rise regime considered below.

The local transverse rise structure is formulated pointwise.
To obtain geometric units on $\mathcal M$ that support a coherent local escape geometry, we collect together regions on which the neutral tangential
structure and the transverse rise law remain uniform.
This leads to the notion of an escape component.

\begin{definition}[Escape component]
\label{def:escape_component}
An escape component is a measurable set $\Com\subset\mathcal M$ with $\mu(\Com)>0$ such that:
\begin{enumerate}
\item[\textnormal{(i)}] $\Com$ admits a uniform local transverse rise structure;

\item[\textnormal{(ii)}] $\Com$ is $\mu$-essentially connected;

\item[\textnormal{(iii)}] $\Com$ is $\subseteq_\mu$-maximal among measurable sets with properties
\textnormal{(i)}--\textnormal{(ii)}.
\end{enumerate}
\end{definition}

Condition \textnormal{(i)} requires that, for $\mathcal L^1$-a.e.\ $\delta>0$, the slice
$\Com\cap\phi^{-1}(\delta)$ admit a finite covering by adapted rise charts $\{\chi_\delta^{(j)}\}_j$
whose images are contained in $\Com$ up to a $\mu$-null set, and for which the associated local quantities
can be chosen uniformly across the covering.
These quantities include the common $\mu$-a.e.\ value of the flat dimension on $\Com$, the transverse rise
exponent, the transverse sector, the chart radii, the rise constants, the density bounds, and the bi-Lipschitz constants.
After shrinking the chart neighborhoods if necessary, this covering exhausts, up to a $\mu$-null set, the
contribution of $\Com$ to sufficiently thin shells above level $\delta$.

Because $\mu$ is finite, the family of $\mu$-relevant escape components is at most countable.
By maximality, distinct escape components are pairwise disjoint modulo $\mu$-null sets.
On the regular shell-generating part considered in the sequel, we assume that the $\mu$-relevant escape
components cover $\mathcal M$ modulo $\mu$.
We then fix an indexing
$\{\Com_\lambda\}_{\lambda\in\Lambda}$
such that
\[
\mathcal M=\bigcup_{\lambda\in\Lambda}\Com_\lambda
\qquad\text{mod }\mu .
\]
For each $\lambda\in\Lambda$, let $d_\lambda$ denote the common $\mu$-a.e.\ value of $d_{\mathrm{fl}}$ on
$\Com_\lambda$, and let $\beta_\lambda$ denote the transverse rise exponent associated with
$\Com_\lambda$.
We write
\[
\mathfrak e_\lambda
:=
\mathcal L^{m-d_\lambda}\bigl(\mathbb E^{m-d_\lambda}\bigr)
\]
for the common $(m-d_\lambda)$-dimensional volume of the unit transverse sector
$\mathbb E^{m-d_\lambda}$ used in the uniform adapted charts of $\Com_\lambda$.

%%%%%%%%%%%%%%%%%%%%%%%%%%%%%%%%%%%%%%%%%%%%%%%%%%%%%%%%%%%%%%%%%%%%

\section{Shell Growth on the Energy Axis}
\label{sec:shell_energy_axis}

Shell growth is the energy-axis trace of shell geometry on $\mathcal M$.
As the threshold increases from $\delta$ to $\delta+t$, the escape-set increment
$X_{\delta+t}\setminus X_\delta$ is generated from component shell bases through transverse opening.
The transverse rise law controls both the emergence of vulnerable mass and the finite-scale shape of
the generated thin shell.
Thus energy-axis growth is a componentwise projection of shell generation on the reference manifold.

\subsection{Shell Generation Geometry}
\label{subsec:shell_side_generation_geometry}

For $\delta>0$, define the energy shell of the escape-energy field at level $\delta$ by
\[
\mathit{\Sigma}_\delta:=\phi^{-1}(\delta).
\]
This is the level interface of the escape-set family $\{X_\delta\}_{\delta>0}$, namely the locus from which new vulnerable mass is generated when the energy threshold is raised.
For $t>0$, define the corresponding thin shell by
\[
\mathrm{Th}_{\delta,t}
:=
\{x\in\mathcal M:\delta<\phi(x)\le \delta+t\}
=
X_{\delta+t}\setminus X_\delta .
\]
Thus the shell is the level interface at energy $\delta$, while the thin shell is the finite-thickness
region in which the resulting growth becomes visible.

The thin shell $\mathrm{Th}_{\delta,t}$ is generated from the shell through the escape components that meet
$\mathit{\Sigma}_\delta$.
For a fixed component $\Com_\lambda$, the shell locus at energy $\delta$ is the intrinsic set
\[
\mathit{\Sigma}_{\delta,\lambda}:=\mathit{\Sigma}_\delta\cap\Com_\lambda .
\]
It is the part of the energy shell carried by $\Com_\lambda$, and hence the intrinsic shell base for
the increment generated by that component.
Adapted rise charts resolve this base locally and provide the transverse coordinates in which the
escape-set increment opens.

The local transverse rise structure then converts this shell base into a finite-thickness contribution.
For fixed $\delta$ and $\Com_\lambda$, the thin shell generated by $\Com_\lambda$ is obtained, up to
$\mu$-null sets, by thickening the same shell base along the adapted transverse opening coordinates, with $t$
setting the transverse opening scale.

\begin{proposition}[Fixed-base generation]
\label{prop:fixed_base_generation}
Fix $\delta>0$ and $\lambda\in\Lambda$.
Assume that the shell locus $\mathit{\Sigma}_{\delta,\lambda}$ admits a finite covering by adapted rise charts
$
\{
\chi_j:S_j\times \mathbb E_{r_j}^{m-d_\lambda}\to\mathcal M
\}_{j=1}^N
$
that exhaust the generated thin shell $\mathrm{Th}_{\delta,t}\cap\Com_\lambda$ modulo $\mu$ for all sufficiently small $t>0$.
Then there exist $t^\ast>0$ and positive functions
$\underline r\, ,\,\overline r$ in $t$ such that for every $t\in(0,t^\ast)$,
\[
\bigcup_{j=1}^N
\chi_j\bigl(S_j\times \mathbb E_{\underline r(t)}^{m-d_\lambda}\bigr)
\subseteq_\mu
\mathrm{Th}_{\delta,t} \cap \Com_\lambda
\subseteq_\mu
\bigcup_{j=1}^N
\chi_j\bigl(S_j\times \mathbb E_{\overline r(t)}^{m-d_\lambda}\bigr).
\]
\end{proposition}

\begin{proof}
Set $\beta:=\beta_\lambda$, $\underline c:=\underline c_\lambda$, $\overline c:=\overline c_\lambda$, and
$r_*:=\min_{1\le j\le N}r_j>0$.
Write $U_j:=\operatorname{Im}(\chi_j)$.
By the thin-shell exhaustion, there exists $t^\ast>0$ such that
\[
\mathrm{Th}_{\delta,t}\cap\Com_\lambda
\subseteq_\mu
\bigcup_{j=1}^N U_j
\qquad
\text{for all }t\in(0,t^\ast).
\]
After decreasing $t^\ast$ if necessary, assume $\underline c\,r_*^\beta>t^\ast$.
For $t\in(0,t^\ast)$ define
\[
\underline r(t):=\overline c^{-1/\beta}t^{1/\beta},
\qquad
\overline r(t):=\underline c^{-1/\beta}t^{1/\beta}.
\]
Then $\overline r(t)<r_*$, and
\[
\|z\|\le \underline r(t)
\quad\Longrightarrow\quad
\overline c\,\|z\|^\beta\le t,
\qquad
\underline c\,\|z\|^\beta\le t
\quad\Longrightarrow\quad
\|z\|\le \overline r(t).
\tag{$*$}
\]

We first prove the inner inclusion.
Fix $j$ and take
\[
x=\chi_j(y,z)\in
\chi_j\bigl(S_j\times\mathbb E_{\underline r(t)}^{m-d_\lambda}\bigr)
\]
outside the chartwise exceptional set.
Since $\chi_j$ is adapted at energy $\delta$, one has $\phi(\chi_j(y,0))=\delta$.
The rise bound gives
\[
\underline c\,\|z\|^\beta
\le
\phi(\chi_j(y,z))-\delta
\le
\overline c\,\|z\|^\beta .
\]
If $z\neq0$, then ($*$) gives $0<\phi(x)-\delta\le t$, hence $x\in\mathrm{Th}_{\delta,t}$.
The zero slice $S_j\times\{0\}$ is
$(\mathcal L^{d_\lambda}\otimes\mathcal L^{m-d_\lambda})$-null, and its image is $\mu$-null by the chart density bound.
Since the chart images are contained in $\Com_\lambda$ modulo $\mu$, it follows that
\[
\bigcup_{j=1}^N
\chi_j\bigl(S_j\times\mathbb E_{\underline r(t)}^{m-d_\lambda}\bigr)
\subseteq_\mu
\mathrm{Th}_{\delta,t}\cap\Com_\lambda .
\]

We next prove the outer inclusion.
Take $x\in \mathrm{Th}_{\delta,t}\cap\Com_\lambda$ outside the $\mu$-null exceptional set.
By exhaustion, $x=\chi_j(y,z)$ for some $j$, with $y\in S_j$ and
$z\in\mathbb E_{r_j}^{m-d_\lambda}$.
Since $x\in\mathrm{Th}_{\delta,t}$, one has $0<\phi(x)-\delta\le t$.
Using $\phi(\chi_j(y,0))=\delta$ and the lower rise bound,
\[
\underline c\,\|z\|^\beta
\le
\phi(\chi_j(y,z))-\delta
=
\phi(x)-\delta
\le t .
\]
Thus $\|z\|\le\overline r(t)$ by ($*$).

It remains to convert this norm bound into membership in the smaller transverse sector.
The boundary case $\|z\|=\overline r(t)$ lies in a transverse sphere and is coordinate-null, hence may be discarded modulo $\mu$.
Thus we may assume $\|z\|<\overline r(t)$.
Since $z\in\mathbb E_{r_j}^{m-d_\lambda}$, write $z=r_j e$ with $e\in\mathbb E^{m-d_\lambda}$.
Because $\overline r(t)<r_j$ and $\|z/\overline r(t)\|<1$, the unit-truncated cone property gives
\[
\frac{z}{\overline r(t)}\in\mathbb E^{m-d_\lambda}.
\]
Hence $z\in\mathbb E_{\overline r(t)}^{m-d_\lambda}$, and therefore
\[
x\in
\chi_j\bigl(S_j\times\mathbb E_{\overline r(t)}^{m-d_\lambda}\bigr).
\]
Consequently,
\[
\mathrm{Th}_{\delta,t}\cap\Com_\lambda
\subseteq_\mu
\bigcup_{j=1}^N
\chi_j\bigl(S_j\times\mathbb E_{\overline r(t)}^{m-d_\lambda}\bigr).
\]
Combining the two inclusions proves the claim.
\end{proof}

Proposition~\ref{prop:fixed_base_generation} makes the source of the thin-shell increment intrinsic.
For fixed $\delta$ and $\Com_\lambda$, the set
$\mathrm{Th}_{\delta,t}\cap\Com_\lambda$ is generated from the fixed shell base
$\mathit{\Sigma}_{\delta,\lambda}\subset\mathit{\Sigma}_\delta$.
The parameter $t$ changes only the transverse opening scale; it does not change the base from which the
increment is produced.
Hence the componentwise thin shells form a nested transverse thickening over a fixed shell locus.

This fixed-base structure separates two shell growth types.
If $d_\lambda=m-1$, the base is hypersurface-like, and the increment is the continuation of an existing
shell face.
If $d_\lambda<m-1$, the base is lower-dimensional, and the increment is created by transverse expansion
away from that locus.
The dimension of the shell base determines whether the local regime is shell-face propagation or codimensional birth.

\subsection{Mass Increment and Energy-Axis Projection}
\label{subsec:mass_increment_energy_projection}

Shell generation is accompanied by a definite growth of vulnerable mass.
For $\delta>0$ and $t>0$, write
$\Delta_\delta(t):=\mu(\mathrm{Th}_{\delta,t})$
for the total $\mu$-measure increment of the escape set, and write
$\Delta_{\delta,\lambda}(t):=
\mu(\mathrm{Th}_{\delta,t}\cap\Com_\lambda)$
for the corresponding componentwise $\mu$-measure increment carried by $\Com_\lambda$.
Within the regular shell-generating regime, since $\{\Com_\lambda\}_{\lambda\in\Lambda}$ partitions
$\mathcal M$ modulo $\mu$-null sets,
\[
\Delta_\delta(t)=\sum_{\lambda\in\Lambda}\Delta_{\delta,\lambda}(t).
\]
For fixed $\delta$ and $\lambda$, the map
$t\mapsto\Delta_{\delta,\lambda}(t)$ is monotone nondecreasing.
Thus a component whose contribution is visible at arbitrarily small thickness is visible throughout the
thin-shell regime.
We call $\Com_\lambda$ thin-shell active at energy $\delta$ if
\[
\Delta_{\delta,\lambda}(t)>0
\qquad
\text{for all sufficiently small }t>0.
\]
For fixed $\delta$ and $\Com_\lambda$, componentwise thin-shell mass is produced by transverse thickening along the same shell base.
The chart density bounds convert this thickening into a quantitative increment law.

\begin{theorem}[Componentwise increment law]
\label{thm:componentwise_increment_law}
Let $J\subset(0,\infty)$ be an open interval.
Then for $\mathcal L^1$-a.e.\ $\delta\in J$ and every $\lambda\in\Lambda$ such that $\Com_\lambda$
is thin-shell active at energy $\delta$, there exist $t_{\delta,\lambda}>0$ and constants
$\underline M_{\delta,\lambda},\overline M_{\delta,\lambda}>0$ such that
\[
\underline\kappa_\lambda\,\underline M_{\delta,\lambda}\,t^{\gamma_\lambda}
\le
\Delta_{\delta,\lambda}(t)
\le
\overline\kappa_\lambda\,\overline M_{\delta,\lambda}\,t^{\gamma_\lambda}
\qquad
\text{for all }t\in(0,t_{\delta,\lambda}),
\]
where
\[
\gamma_\lambda:=\frac{m-d_\lambda}{\beta_\lambda},
\qquad
\underline\kappa_\lambda
:=
\underline w_{\lambda}\,\mathfrak e_\lambda\,(\overline c_\lambda)^{-\gamma_\lambda},
\qquad
\overline\kappa_\lambda
:=
\overline w_{\lambda}\,\mathfrak e_\lambda\,(\underline c_\lambda)^{-\gamma_\lambda}.
\]
% Here $\underline M_{\delta,\lambda}$ and $\overline M_{\delta,\lambda}$ encode the size of the shell
% slice $\Com_\lambda\cap\phi^{-1}(\delta)$ carried by $\Com_\lambda$ at energy $\delta$.
\end{theorem}

\begin{proof}
Since the family of $\mu$-relevant escape components is at most countable, we may restrict to a full
$\mathcal L^1$-measure subset of $J$ on which the uniform local transverse rise structure holds for
every relevant component.
Fix such a $\delta\in J$, and let $\Com_\lambda$ be thin-shell active at energy $\delta$.

Set $\beta:=\beta_\lambda$ and
$\gamma_\lambda=\frac{m-d_\lambda}{\beta}$.
Let
\[
\{
\chi_j:S_j\times \mathbb E_{r_j}^{m-d_\lambda}\to\mathcal M
\}_{j=1}^N
\]
be a finite adapted covering of $\mathit{\Sigma}_{\delta,\lambda}$ with the corresponding thin-shell
exhaustion.
The uniform rise constants and density bounds on this covering are
$\underline c_\lambda,\overline c_\lambda,\underline w_\lambda,\overline w_\lambda$.
Thin-shell activity and exhaustion imply that the adapted covering is nonempty.

By Proposition~\ref{prop:fixed_base_generation}, there exists $t^\ast>0$ such that for all
$t\in(0,t^\ast)$, with
\[
\underline r(t):=\overline c_\lambda^{-1/\beta}t^{1/\beta},
\qquad
\overline r(t):=\underline c_\lambda^{-1/\beta}t^{1/\beta},
\]
the sets
\[
\underline U_j(t):=\chi_j\bigl(S_j\times\mathbb E_{\underline r(t)}^{m-d_\lambda}\bigr),
\qquad
\overline U_j(t):=\chi_j\bigl(S_j\times\mathbb E_{\overline r(t)}^{m-d_\lambda}\bigr)
\]
satisfy
\[
\bigcup_{j=1}^N\underline U_j(t)
\subseteq_\mu
\mathrm{Th}_{\delta,t}\cap\Com_\lambda
\subseteq_\mu
\bigcup_{j=1}^N\overline U_j(t).
\]

For each $j$, the coordinate density bounds give
\[
\mu(\underline U_j(t))
\ge
\underline w_\lambda\,\mathcal L^{d_\lambda}(S_j)\,
\mathfrak e_\lambda\,\underline r(t)^{m-d_\lambda}
=
\underline\kappa_\lambda\,\mathcal L^{d_\lambda}(S_j)t^{\gamma_\lambda},
\]
and
\[
\mu(\overline U_j(t))
\le
\overline w_\lambda\,\mathcal L^{d_\lambda}(S_j)\,
\mathfrak e_\lambda\,\overline r(t)^{m-d_\lambda}
=
\overline\kappa_\lambda\,\mathcal L^{d_\lambda}(S_j)t^{\gamma_\lambda}.
\]
The first inclusion gives
\[
\Delta_{\delta,\lambda}(t)
\ge
\max_{1\le j\le N}\mu(\underline U_j(t)),
\]
while the second inclusion and subadditivity give
\[
\Delta_{\delta,\lambda}(t)
\le
\sum_{j=1}^N\mu(\overline U_j(t)).
\]
Therefore
\[
\Delta_{\delta,\lambda}(t)
\ge
\underline\kappa_\lambda
\Bigl(\max_{1\le j\le N}\mathcal L^{d_\lambda}(S_j)\Bigr)t^{\gamma_\lambda},
\]
and
\[
\Delta_{\delta,\lambda}(t)
\le
\overline\kappa_\lambda
\Bigl(\sum_{j=1}^N\mathcal L^{d_\lambda}(S_j)\Bigr)t^{\gamma_\lambda}.
\]
Set
\[
\underline M_{\delta,\lambda}:=\max_{1\le j\le N}\mathcal L^{d_\lambda}(S_j),
\qquad
\overline M_{\delta,\lambda}:=\sum_{j=1}^N\mathcal L^{d_\lambda}(S_j).
\]
These constants are positive and finite, and the claim follows with
$t_{\delta,\lambda}:=t^\ast$.
\end{proof}

Theorem~\ref{thm:componentwise_increment_law} identifies the measure trace of fixed-base generation.
For a fixed component, the onset of vulnerable mass near energy $\delta$ is governed by the same
shell geometry that generates the thin shell.
At sufficiently small thickness, this onset is confined to a single asymptotic mode: the power
$t^{\gamma_\lambda}$ is fixed by the component geometry, while the coefficients record the size and
density of the shell base.
The growth of $\mu$ along the energy axis is therefore the projected measure effect of transverse
opening over a shell base, and remains an intrinsic trace of the local escape geometry on
$\mathcal M$.
Different components may contribute different local growth laws because they realize different
dimensional types of shell generation.

The componentwise increment law admits an energy-axis projection.
Let $\eta:=\phi_\#\mu$ be the pushforward of $\mu$ under $\phi$.
Then, for every $\delta>0$ and $t>0$,
\[
\eta((\delta,\delta+t])
=
\mu(\phi^{-1}((\delta,\delta+t]))
=
\mu(\mathrm{Th}_{\delta,t}).
\]
Thus thin-shell $\mu$-mass is the projected mass of a one-dimensional energy interval.
For $\delta>0$, set
\[
\Lambda_{\mathrm{act}}(\delta)
:=
\{\lambda\in\Lambda:\ \Com_\lambda \text{ is thin-shell active at energy }\delta\},
\qquad
\Gamma_{\mathrm{act}}(\delta)
:=
\{\gamma_\lambda:\lambda\in\Lambda_{\mathrm{act}}(\delta)\}.
\]
The set $\Gamma_{\mathrm{act}}(\delta)$ is the active exponent spectrum at energy $\delta$.
It retains only the growth exponents carried by components that actually contribute to the thin shell at that level.

\begin{assumption}[Finite comparable active spectrum]
\label{ass:finite_comparable_active_spectrum}
Let $J\subset(0,\infty)$ be an open interval, and let $I\Subset J$ be an interval.
For $\mathcal L^1$-a.e.\ $\delta\in I$:
\begin{enumerate}
\item[\textnormal{(i)}] 
The active component set $\Lambda_{\mathrm{act}}(\delta)$ is finite. In particular,
$\Gamma_{\mathrm{act}}(\delta)$ is finite.

\item[\textnormal{(ii)}] 
Uniformly at sufficiently small thickness on $I$, $
\Delta_\delta(t)
=
\sum_{\lambda\in\Lambda_{\mathrm{act}}(\delta)}
\Delta_{\delta,\lambda}(t)$.
\end{enumerate}
Moreover, there exists a measurable function
\[
W:I\times(0,\infty)\to[0,\infty)
\]
such that for $\mathcal L^1$-a.e.\ $\delta\in I$:
\begin{enumerate}
\item[\textnormal{(iii)}] $W(\delta,\gamma)=0$ whenever $\gamma\notin\Gamma_{\mathrm{act}}(\delta)$;

\item[\textnormal{(iv)}] for each $\gamma\in\Gamma_{\mathrm{act}}(\delta)$,  $W(\delta,\gamma)$ is uniformly comparable on $I$ to the total coefficient of the active components with exponent $\gamma$ in the componentwise increment law.
\end{enumerate}
\end{assumption}

Conditions \textnormal{(i)} and \textnormal{(ii)} ensure that, at almost every energy in $I$, the active components give a finite and exhaustive description of the sufficiently thin shell on a common small-thickness range.
The remaining conditions assign a projected coefficient to each visible growth order.
The function $W$ is an interval-level projected mass coefficient, indexed by energy $\delta$ and growth exponent $\gamma$.
Condition \textnormal{(iii)} fixes its support on $\Gamma_{\mathrm{act}}(\delta)$, while Condition \textnormal{(iv)} makes it uniformly comparable, on $I$, to the corresponding sum of componentwise coefficients in Theorem~\ref{thm:componentwise_increment_law}.
This weighted spectrum is sufficient to recover the vulnerable mass of an interval from the active growth orders along that interval.

\begin{corollary}[Interval spectral recovery]
\label{cor:interval_spectral_recovery}
Let $J\subset(0,\infty)$ be an open interval, let $a<b$ with $[a,b]\Subset J$, and set
$I=(a,b]$.
Suppose that $I$ is equipped with an interval-level projected mass coefficient $W$.
Then the projected mass of $I$ satisfies
\[
\eta(I)
\asymp
\lim_{t\downarrow0}
\int_I
\Bigl(
\sum_{\gamma\in\Gamma_{\mathrm{act}}(\delta)}
W(\delta,\gamma)\,t^{\gamma-1}
\Bigr)d\delta,
\]
whenever the limit on the right exists as a finite quantity.
\end{corollary}

\begin{proof}
By Assumption~\ref{ass:finite_comparable_active_spectrum} and
Theorem~\ref{thm:componentwise_increment_law}, after discarding an
$\mathcal L^1$-null subset of $I$, the active components exhaust the thin shell on a common
small-thickness range, and the componentwise increment laws may be grouped by their common exponent.
Thus, for all sufficiently small $t>0$,
\[
\Delta_\delta(t)
\asymp
\sum_{\gamma\in\Gamma_{\mathrm{act}}(\delta)}
W(\delta,\gamma)t^\gamma
\]
for $\mathcal L^1$-a.e.\ $\delta\in I$, with comparison constants independent of
$\delta$ and $t$ in this common range.
Since
\[
\Delta_\delta(t)=\eta((\delta,\delta+t]),
\]
division by $t$ and integration over $I$ give
\[
\frac1t\int_I \eta((\delta,\delta+t])\,d\delta
\asymp
\int_I
\Bigl(
\sum_{\gamma\in\Gamma_{\mathrm{act}}(\delta)}
W(\delta,\gamma)t^{\gamma-1}
\Bigr)d\delta .
\]

Define
\[
K_{I,t}(u):=\frac1t\,\big|I\cap(u-t,u]\big|,
\qquad u\in(0,\infty).
\]
By Tonelli's theorem,
\[
\frac1t\int_I \eta((\delta,\delta+t])\,d\delta
=
\int_{(0,\infty)}K_{I,t}(u)\,d\eta(u).
\]
Since $I=(a,b]$, one has $K_{I,t}(u)\to\mathbf 1_I(u)$ for every
$u\in(0,\infty)$, and $0\le K_{I,t}\le1$.
As $\eta$ is finite, dominated convergence yields
\[
\lim_{t\downarrow0}
\frac1t\int_I \eta((\delta,\delta+t])\,d\delta
=
\eta(I).
\]
Taking the limit in the preceding comparison gives the claimed interval recovery, whenever the
right-hand limit exists as a finite quantity.
\end{proof}

Corollary~\ref{cor:interval_spectral_recovery} gives the interval-scale form of the preceding
componentwise law.
After projection to the energy axis, componentwise contributions are no longer resolved by their
individual carriers, but by the active growth exponents visible at each energy level.
The projected coefficient $W$ records the mass weight attached to each visible order, and integration
along $I$ accumulates these weighted orders into the interval mass $\eta(I)$.
Since the spectral contributions are nonnegative, Tonelli's theorem
allows the exponent-wise summation and the integration in $\delta$ to be interchanged.
Thus the interval mass may be read, up to the fixed comparability of the projected coefficients, as an
accumulation over energy levels or as a sum of exponent-resolved layers gathered along the interval.

This also separates the present recovery from the coarea formula.
The coarea formula decomposes $\mu$ by slicing along the level sets of
$\phi$, with the Jacobian $J_{\mathcal M}\phi$ controlling the regular infinitesimal contribution of
each shell.
The corollary concerns the mass produced when a shell base opens transversely into a thin shell.
Such mass need not be carried by the shell itself as $\mu$-mass.
It appears through the finite-thickness opening governed by the active component geometry.
Consequently, the relevant data are not only the level shell $\mathit{\Sigma}_\delta$, but the
componentwise generation laws carried by the active shell bases.
Coarea resolves level-set slicing.
The corollary resolves the spectral organization of interval mass produced by shell generation.

%%%%%%%%%%%%%%%%%%

\subsection{Normalized Occupancy}
\label{subsec:normalized_occupancy}

At positive thickness, shell generation produces an $m$-dimensional thin-shell region inside
$\mathcal M$.
Finite-scale morphology records the local shape of this generated region, beyond dimension and mass
order.
Here it is encoded through local measure distributions, which make thin-shell shape comparable at a
fixed observation scale.
A mild rescaling stability of the adapted rise structure supplies the regularity needed for such
comparisons to stabilize under rescaling.

\begin{assumption}[Asymptotically stationary]
\label{ass:asymptotically_stationary_local_rise}
Fix $\delta>0$.
For every point $x\in\phi^{-1}(\delta)$ lying on an active shell patch carried by a component
$\Com_\lambda$, there exists an adapted rise chart at energy $\delta$ through $x$,
\[
\chi:S\times \mathbb E_{r_\chi}^{m-d_\lambda}\to\mathcal M,
\qquad
x=\chi(y,0),
\]
where $y$ is a Lebesgue density point of $S$ and $w$ is the coordinate density of $\mu$ in $\chi$,
together with a function
$g:\mathbb S^{m-d_\lambda-1}\to(0,\infty)$ bounded above and below away from $0$, and a constant
$w_0>0$, such that as $\varepsilon\downarrow0$:
\begin{enumerate}
\item[\textnormal{(i)}] \textnormal{Stationary transverse rise.}
\[
\sup_{\substack{u\in S\cap \mathbb B_\varepsilon^{d_\lambda}(y)\\
z\in \mathbb E_\varepsilon^{m-d_\lambda},\ z\neq0}}
\left|
\frac{\phi(\chi(u,z))-\delta}{\|z\|^{\beta_\lambda}}
-
g\!\left(\frac{z}{\|z\|}\right)
\right|
\longrightarrow0.
\]

\item[\textnormal{(ii)}] \textnormal{Stationary coordinate density.}
\[
\sup_{\substack{u\in S\cap \mathbb B_\varepsilon^{d_\lambda}(y)\\
z\in \mathbb E_\varepsilon^{m-d_\lambda}}}
\bigl|w(u,z)-w_0\bigr|
\longrightarrow0.
\]
\end{enumerate}
\end{assumption}

Assumption~\ref{ass:asymptotically_stationary_local_rise} upgrades the adapted rise chart at $x$ from
coarse rise bounds to a stable rescaling.
Condition \textnormal{(i)} fixes the limiting transverse opening profile, and Condition
\textnormal{(ii)} fixes the limiting chart density of $\mu$.
Consequently, both the transverse geometry and the local measure weight admit well-defined rescaled
limits near the active shell point.

Finite-scale shape is made comparable by normalization.
Each local observation window is mapped to a fixed bounded convex template $K\subset\mathbb R^m$, and the newly vulnerable set is recorded by the pushed-forward $\mu$-measure it induces on that template. Its total mass gives the occupied fraction of the window, while its distribution on $K$ records the finite-scale shape of the occupied region.

\begin{definition}[Affine normalized window]
\label{def:admissible_affine_normalized_window}
Fix $c_{\mathrm{aff}}\ge1$ and a bounded convex set
$K\subset\mathbb R^m$ with $0\in\operatorname{int}(K)$.
Let $x\in\mathcal M$ and $r>0$.
An affine $K$-normalized local window at $(x,r)$ is a measurable set
$\mathcal W_r\subset\mathcal M$ with $x\in\mathcal W_r$, equipped with a measurable normalization map
$N_r:\mathcal W_r\to K$, such that in a coordinate chart on $\mathcal M$ containing $\mathcal W_r$,
\[
N_r(u)=A_ru+b_r,
\qquad
A_r\in \operatorname{GL}(m,\mathbb{R}),\quad b_r\in\mathbb R^m,
\]
where
$\|A_r\|\le c_{\mathrm{aff}}r^{-1}$ and $\|A_r^{-1}\|\le c_{\mathrm{aff}}r $.
\end{definition}

The window $\mathcal W_r$ is the local observation region at scale $r$, and $N_r$ places this region on
the fixed template $K$.
The bounds on $A_r$ and $A_r^{-1}$ make this placement comparable to an $r^{-1}$ rescaling, with
uniformly controlled distortion.
Thus windows observed at different scales can be compared on a common reference domain, and differences
in thin-shell shape are read from the occupied region on $K$ rather than from the original size or
position of the window.
The affine form is used to eliminate translation, orientation, and scale without allowing the normalization to absorb finite-scale features of the generated thin shell.
Compared with a nonlinear normalization, it avoids mixing the intrinsic morphology of shell growth with
coordinate artifacts, and hence leaves the finite-scale shape visible up to controlled linear deformation.

After normalization, the local thin-shell increment is represented by its induced measure on the common
template $K$.
For $x\in\phi^{-1}(\delta)$, an affine $K$-normalized local window $\mathcal W_r$ at
$(x,r)$ with normalization map $N_r$, and $t>0$, write
\[
\omega_{r,t}
:=
\frac{(N_r)_\#\bigl(\mu\!\lfloor_{\mathrm{Th}_{\delta,t}\cap \mathcal W_r}\bigr)}{\mu(\mathcal W_r)}.
\]
The mass $\omega_{r,t}(K)$ gives the occupied fraction of the window, while the distribution of
$\omega_{r,t}$ on $K$ gives a finite-scale shape profile of the occupied region.
When $x$ lies on an active shell patch carried by $\Com_\lambda$, this profile depends on two scales:
the window scale $r$ fixes the observation scale, while the energy increment $t$ produces transverse
opening at scale $t^{1/\beta_\lambda}$.
Since the object is the finite-scale shape profile rather than only the amount of mass, a nondegenerate
limit requires the opening scale and the observation scale to shrink in step.
This singles out the matched ratio
$h:=t^{1/\beta_\lambda}/r$,
which measures transverse opening in units of observation scale.

Normalized profiles are finite measures on the common template $K$.
For two finite measures $\nu_1$ and $\nu_2$ on $K$, their finite-scale shapes are distinguished using the bounded-Lipschitz distance
\[
d_{\mathrm{BL}}(\nu_1,\nu_2)
:=
\sup\Bigl\{
\Bigl|\int_K f\,d(\nu_1-\nu_2)\Bigr|:
f:K\to\mathbb R,\ \|f\|_{L^\infty(K)}\le 1,\ \mathrm{Lip}(f)\le 1
\Bigr\}.
\]
This distance is sensitive both to the occupied mass and to its spatial distribution on $K$.
With this notion of profile separation, finite-scale morphology can be formulated as a stability statement for normalized occupancy profiles.

\begin{theorem}[Matched-scale occupancy law]
\label{thm:matched_scale_occupancy_law}
Let $\delta>0$, and let $x\in\phi^{-1}(\delta)$ lie on an active shell patch carried by
$\Com_\lambda$.
Fix a bounded convex set $K\subset\mathbb R^m$ with $0\in\operatorname{int}(K)$, and a bounded
matched-ratio interval $J\subset[0,\infty)$ with $0\in J$.
Then there exists a family of finite measures $\{\nu_h\}_{h\in J}$ on $K$ with the following property.

For every compact interval $I\Subset J\setminus\{0\}$, there exist $t_I>0$ and a function
\[
\alpha_I:(0,t_I]\to[0,\infty),
\qquad
\alpha_I(t)\to0 \quad\text{as }t\downarrow0,
\]
such that, for every $h\in I$ and $0<t\le t_I$, setting
$r=r(t,h):=\frac{t^{1/\beta_\lambda}}{h}$,
and taking $\mathcal W_r$ to be the affine $K$-normalized local window at $(x,r)$ associated with the
adapted rise chart at $x$, with normalization map $N_r$, one has
\[
d_{\mathrm{BL}}\!\bigl(\omega_{r,t},\nu_h\bigr)\le \alpha_I(t).
\]
Moreover, there exists $h_0>0$ such that
\[
\nu_h(K)\asymp h^{m-d_\lambda}
\qquad
\text{for all }h\in J\cap(0,h_0].
\]
\end{theorem}

\begin{proof}
Write $\beta:=\beta_\lambda$.
Use the splitting
\[
\mathbb R^m=\mathbb R^{d_\lambda}\times\mathbb R^{m-d_\lambda}
\]
induced by the adapted chart, and write points of $K$ as $(\xi,\zeta)$.
Let
\[
\chi:S\times\mathbb E_{r_\chi}^{m-d_\lambda}\to\mathcal M,
\qquad
x=\chi(y,0),
\]
together with $g:\mathbb S^{m-d_\lambda-1}\to(0,\infty)$, the chart density $w$, and the limit density
$w_0>0$, be given by Assumption~\ref{ass:asymptotically_stationary_local_rise}.

First define the limiting normalized support.
Let
\[
\mathcal E_\chi^{m-d_\lambda}
:=
\bigcup_{\rho>0}\mathbb E_{\rho}^{m-d_\lambda}
\]
be the transverse opening cone generated by the sector of $\Com_\lambda$, and set
\[
K_\chi
:=
K\cap\bigl(\mathbb R^{d_\lambda}\times\mathcal E_\chi^{m-d_\lambda}\bigr).
\]
Since $K$ is bounded, convex, and contains $0$ in its interior, while the transverse sector has positive
$(m-d_\lambda)$-dimensional measure, one has $\mathcal L^m(K_\chi)>0$.

Define the limiting transverse rise profile by
\[
G(\zeta):=
\begin{cases}
g(\zeta/\|\zeta\|)\|\zeta\|^\beta, & \zeta\neq0,\\
0, & \zeta=0.
\end{cases}
\]
For $h\in J$, set
\[
\nu_h:=
\frac{
\mathbf 1_{K_\chi}\mathbf 1_{\{0<G(\zeta)\le h^\beta\}}\,
\mathcal L^m\!\lfloor_K
}{
\mathcal L^m(K_\chi)
}.
\]
This is the candidate limiting occupancy profile.

Fix $I\Subset J\setminus\{0\}$ and put $h_-:=\inf I>0$.
Choose $r_\ast>0$ so small that the rescaled transverse part of $K_\chi$ remains in
$\mathbb E_{r_\chi}^{m-d_\lambda}$ whenever $0<r\le r_\ast$.
For such $r$, define
\[
K_r
:=
\{(\xi,\zeta)\in K:\ y+r\xi\in S,\ r\zeta\in\mathbb E_{r_\chi}^{m-d_\lambda}\},
\]
and take the chart-induced normalized window
\[
\mathcal W_r
:=
\chi\bigl(\{(y+r\xi,r\zeta):(\xi,\zeta)\in K_r\}\bigr),
\qquad
N_r\bigl(\chi(y+r\xi,r\zeta)\bigr):=(\xi,\zeta).
\]
The tangential condition converges in measure because $y$ is a Lebesgue density point of $S$.
The transverse condition converges to
$\zeta\in\mathcal E_\chi^{m-d_\lambda}$ by the radial structure of the transverse sector.
Hence
\[
\mathcal L^m(K_r\triangle K_\chi)\to0
\qquad
\text{as }r\downarrow0 .
\]
Since $K$ is bounded, this follows by applying the Lebesgue density theorem on bounded
rescaled windows and dominated convergence on the fixed template $K$.

Set $t_I:=(h_-r_\ast)^\beta$.
For $h\in I$ and $0<t\le t_I$, put $r=r(t,h):=t^{1/\beta}/h$.
Then $0<r\le r_\ast$ and
\[
\sup_{h\in I}r(t,h)\le t^{1/\beta}/h_-\to0
\qquad
\text{as }t\downarrow0 .
\]

We now reduce the claim to convergence of densities on the fixed template $K$.
For $h\in I$ and $0<t\le t_I$, define the unnormalized window density
\[
B_{t,h}(\xi,\zeta)
:=
\mathbf 1_{K_{r(t,h)}}(\xi,\zeta)\,
w(y+r(t,h)\xi,r(t,h)\zeta),
\]
and the unnormalized occupancy density
\[
H_{t,h}(\xi,\zeta)
:=
B_{t,h}(\xi,\zeta)\,
\mathbf 1_{\{0<\phi(\chi(y+r(t,h)\xi,r(t,h)\zeta))-\delta\le t\}} .
\]
Thus $B_{t,h}$ records the normalized window before imposing the thin-shell condition, while
$H_{t,h}$ records the occupied thin-shell part of that window.

By the pushforward definition of $\omega_{r,t}$ and the normalization identity
$N_r(\chi(y+r\xi,r\zeta))=(\xi,\zeta)$, for every bounded measurable $f$ on $K$,
\[
\int_K f\,d\omega_{r(t,h),t}
=
\frac{\int_K f(\xi,\zeta)H_{t,h}(\xi,\zeta)\,d\xi\,d\zeta}
{\int_K B_{t,h}(\xi,\zeta)\,d\xi\,d\zeta}.
\]
The common Jacobian factor $r^m$ cancels between numerator and denominator.

The two density limits are now identified.
First, from $\mathcal L^m(K_{r(t,h)}\triangle K_\chi)\to0$ and
$w(y+r(t,h)\xi,r(t,h)\zeta)\to w_0$, uniformly for $h\in I$ on the relevant rescaled windows, one gets
\[
\sup_{h\in I}
\left\|
B_{t,h}-w_0\mathbf 1_{K_\chi}
\right\|_{L^1(K)}
\to0 .
\]
Second, along the matched scaling $t=h^\beta r(t,h)^\beta$,
Assumption~\ref{ass:asymptotically_stationary_local_rise} gives
\[
\frac{\phi(\chi(y+r(t,h)\xi,r(t,h)\zeta))-\delta}{r(t,h)^\beta}
\to
G(\zeta)
\]
uniformly for $h\in I$ on the relevant rescaled windows, away from $\zeta=0$.
For each $h>0$, the level set $\{G=h^\beta\}$ has $\mathcal L^m$-measure zero by polar coordinates in
the transverse variable.
Since $I\Subset J\setminus\{0\}$, the transition bands
\[
\{|G-h^\beta|\le\varepsilon\}
\]
have $\mathcal L^m$-measure tending to zero uniformly for $h\in I$ as $\varepsilon\downarrow0$.
Indeed, in polar coordinates in the transverse variable, for each angular direction the band
corresponds to a radial interval whose length is $O(\varepsilon)$ uniformly for $h\in I$,
because $I$ is bounded away from $0$ and $g$ is bounded above and below.
Therefore
\[
\sup_{h\in I}
\left\|
H_{t,h}
-
w_0\,\mathbf 1_{K_\chi}\mathbf 1_{\{0<G(\zeta)\le h^\beta\}}
\right\|_{L^1(K)}
\to0
\qquad
\text{as }t\downarrow0 .
\]

The preceding two convergences imply the convergence of the normalized measures.
Indeed,
\[
\int_K B_{t,h}\,d\mathcal L^m
\to
w_0\,\mathcal L^m(K_\chi)
\]
uniformly for $h\in I$, and this limit is strictly positive.
Hence the measures with densities
\[
\frac{H_{t,h}}{\int_K B_{t,h}\,d\mathcal L^m}
\]
converge in total variation, uniformly for $h\in I$, to the measure with density
\[
\frac{\mathbf 1_{K_\chi}\mathbf 1_{\{0<G(\zeta)\le h^\beta\}}}{\mathcal L^m(K_\chi)} .
\]
This limiting measure is exactly $\nu_h$.
Since bounded-Lipschitz distance is controlled by total variation, it follows that
\[
\sup_{h\in I}
d_{\mathrm{BL}}\!\bigl(\omega_{r(t,h),t},\nu_h\bigr)
\to0
\qquad
\text{as }t\downarrow0 .
\]
Define
\[
\alpha_I(t):=
\sup_{0<s\le t}\ \sup_{h\in I}
d_{\mathrm{BL}}\!\bigl(\omega_{r(s,h),s},\nu_h\bigr),
\qquad
0<t\le t_I .
\]
Then $\alpha_I(t)\to0$, and the asserted stability estimate follows.

It remains to estimate the mass of $\nu_h$ for small $h$.
Let
\[
g^-:=\inf_{\mathbb S^{m-d_\lambda-1}}g,
\qquad
g^+:=\sup_{\mathbb S^{m-d_\lambda-1}}g .
\]
Then $0<g^-\le g^+<\infty$, and for sufficiently small $h>0$,
\[
K_\chi\cap\{\|\zeta\|\le (h^\beta/g^+)^{1/\beta}\}
\subset
K_\chi\cap\{0<G(\zeta)\le h^\beta\}
\subset
K_\chi\cap\{\|\zeta\|\le (h^\beta/g^-)^{1/\beta}\}.
\]
Since $K$ contains a neighborhood of the origin and is bounded, while
$\mathcal E_\chi^{m-d_\lambda}$ has positive transverse opening measure, the two outer volumes are
both comparable to $h^{m-d_\lambda}$ as $h\downarrow0$.
Dividing by $\mathcal L^m(K_\chi)$ gives
\[
\nu_h(K)\asymp h^{m-d_\lambda}.
\]
\end{proof}

Theorem~\ref{thm:matched_scale_occupancy_law} identifies finite-scale morphology through a matched
rescaling law.
It concerns the normalized occupancy profile obtained when the transverse opening scale and the observation scale are coupled through $h=t^{1/\beta_\lambda}/r$.
In this regime, sending $t\downarrow0$ removes absolute scale without erasing normalized shape.
With the chart-induced normalization used in the proof, the measures $\omega_{r(t,h),t}$ converge to a
limiting profile $\nu_h$ on the common template $K$.
Thus the bounded-Lipschitz limit is a shape limit rather than a collapse to a zero object.
The theorem therefore resolves more than an onset exponent, giving a one-parameter family of local
shape profiles indexed by the finite-scale opening ratio $h$.

The normalization $N_r$ used in the proof is the canonical one induced by the adapted rise chart.
It fixes the shell base point and rescales the tangential and transverse chart coordinates by $r^{-1}$.
For another affine normalization $\widetilde N_r$, the relative maps
$T_r:=\widetilde N_r\circ (N_r^0)^{-1}$, with $N_r^0$ denoting the chart-induced normalization,
determine the change of representative.
If $T_r\to T_\infty$ as $r\downarrow0$, the limiting profile becomes $(T_\infty)_\#\nu_h$.
Thus the invariant object is the local occupancy morphology up to controlled affine normalization, and
$\nu_h$ is its chart-induced representative.

The structural content is that this limiting profile is tied to one componentwise shell generation
structure.
Different components may carry different transverse exponents, base dimensions, or limiting rise
profiles, and hence may produce different limiting occupancy families.
Under compatible normalizations, the distance between two matched profiles converges to the distance
between their limiting profiles, not automatically to zero.
Thus finite-scale occupancy distinguishes componentwise shell geometry that is indistinguishable at the
level of vanishing absolute scale.
If arbitrarily small neighborhoods of a point contain several active component geometries, one should
not expect a single affine normalization to make all of them simultaneously nondegenerate.
Local spectral multiplicity can therefore obstruct a uniform finite-scale surrogate for the local
vulnerable set.

\begin{remark}[Local sampling interpretation]
\label{rem:local_sampling_interpretation}
The mass of the normalized occupancy profile also has a local sampling interpretation.
Consider a sufficiently small sampling window $U_r\subset\mathcal M$ centered at $x$, containing the
normalized window $\mathcal W_r$.
Suppose that sampling in $U_r$ is described by a probability measure $\mathbb P$ whose normalized image
on the part of the template $K$ selected by the transverse sector is comparable to a full-dimensional
reference measure, and that, for matched pairs
$(r,t)$, the thin-shell points visible in $U_r$ are carried by $\mathcal W_r$ up to a $\mu$-null set.
In adapted coordinates, the local sampling mass separates, up to fixed comparability constants, into
the tangential presence of the shell base and the transverse occupancy of the generated opening.
Thus the local hit probability of the thin shell has the scale
\[
\mathbb P\bigl(U_r\cap \mathrm{Th}_{\delta,t}\bigr)
\asymp
\Xi(r)\,\nu_h(K),
\qquad
\Xi(r):=
\frac{\mathcal L^{d_\lambda}\!\bigl(S\cap \mathbb B_r^{\,d_\lambda}(y)\bigr)}
{r^{d_\lambda}} .
\]
Here $\Xi(r)$ is the normalized tangential density of the shell base near the coordinate point
$y$, while $\nu_h(K)$ is the normalized transverse occupied fraction inside the generated-opening part
of the template.
At $\mathcal L^{d_\lambda}$-typical shell points, $\Xi(r)\asymp1$, and the relation reduces to
$\mathbb P\bigl(U_r\cap \mathrm{Th}_{\delta,t}\bigr)
\asymp
\nu_h(K)$.
Thus local sampling and normalized occupancy are two readings of the same shell-generated set.
The normalized occupancy measure records its distribution on the generated-opening part of the
template $K$, while the hit probability records its mass under local sampling.
The decay of this probability is therefore the probabilistic trace of the same transverse
codimensional concentration encoded by $\nu_h(K)$.
\end{remark}

%%%%%%%%%%%%%%%%%%%%%%%%%%%%%%%%%%%%%%%%%%%%%%%%%

%%%%%%%%%%%%%%%%%%%%%%%%%%%%%%%%%%%%%%%%%%%%%%%%%

\section{Realized Escape Dynamics}
\label{sec:realized_escape_dynamics}

Realized escape is the ambient dynamics of escape geometry.
The escape-energy field on the reference manifold fixes the scale of vulnerability, but an actual
escape is carried by paths that move through the ambient space.
Along such paths, the observable geometry that generates motion and the ambient extension that measures
remaining escape progress need not coincide.
Realization therefore depends on how local descending motion is converted into reference-side escape
progress, and on whether nearby initial points remain organized by compatible observable histories as
the realized family evolves.

\subsection{Ambient and Observable Geometry}
\label{subsec:ambient_observable_geometry}

The escape-energy field $\phi$ records, on the reference manifold $\mathcal M$, the minimal perturbation required to reach $\partial\Omega$ from an initial point $x\in\mathcal M$.
However, the admissible paths used in the variational specification are not constrained to remain on $\mathcal M$ after they leave the initial point.
Thus, once escape is viewed as an actual realization process, the relevant state is an ambient position $z\in\mathbb R^n$ rather than a manifold point alone.

\begin{definition}[Ambient extension of the escape-energy field]
\label{def:ambient_extension}
The ambient extension of $\phi$ is the function $\hat\phi:\mathbb R^n\to[0,\infty]$ defined by
\[
\hat\phi(z):=\sup_{x\in\mathcal M}\bigl(\phi(x)-p(x,z)\bigr)_+.
\]
\end{definition}

The field $\hat\phi$ is the canonical ambient extension of the manifold escape geometry induced by the same perturbation gauge used in Definition~\ref{def:escape_cost_variational}.
For each initial point $x\in\mathcal M$, the quantity $(\phi(x)-p(x,z))_+$ is the residual lower bound obtained after discounting the escape energy at $x$ by the gauge displacement from $x$ to $z$, and $\hat\phi(z)$ is the supremum of these bounds over all $x\in\mathcal M$.
Thus $\phi$ records required escape energy on the reference manifold, whereas $\hat\phi$ records the corresponding residual required escape energy at ambient positions reached along realized escapes.

\begin{assumption}[Ambient extension regularity]
\label{ass:ambient_extension_regularities}
The ambient extension $\hat\phi$ is finite and locally Lipschitz on $\mathbb R^n$, and $\hat\phi(x)=\phi(x)$ for every $x\in\mathcal M$.
\end{assumption}

Assumption~\ref{ass:ambient_extension_regularities} states that $\hat\phi$ is a legitimate ambient carrier of this same escape geometry.
The identity $\hat\phi|_{\mathcal M}=\phi$ preserves the manifold escape values exactly, so no new geometry is introduced on the reference set.
Local Lipschitz regularity ensures that $\hat\phi$ varies in a controlled manner under ambient displacement and carries a well-defined first-order nonsmooth geometry almost everywhere.
These are therefore consistency requirements on a canonical extension.

The field $\hat\phi$ provides the global ambient escape geometry against which realized escape progress
is measured.
The mechanism itself, however, need not realize escape by directly using this global extension.
Its implementation may depend only on local geometric information available at the ambient positions it
actually reaches.
This motivates the observable field $z\mapsto\psi_z$, which represents the operative local geometry through which the mechanism realizes the next stage of escape from the current position $z$.
In favorable regimes, $\psi_z$ faithfully reflects the local structure of $\hat\phi$ near $z$.
In general, however, the operative geometry carried by $\psi_z$ should be allowed to differ from $\hat\phi$, since the mechanism may rely only on a locally available surrogate geometry, or on a geometry inferred from local probing, rather than on direct access to the true ambient extension.

\begin{definition}[Observable field]
\label{def:observable_geometry_mechanism}
An observable field at $z\in\mathbb R^n$ is a function
\[
\psi_z:U_z\to[0,\infty)
\]
defined on an ambient neighborhood $U_z\subset\mathbb R^n$ of $z$.
\end{definition}

The field $\psi_z$ should be understood as the mechanism's operative local representation of the escape geometry around the current ambient point $z$.
For $u$ near $z$, the value $\psi_z(u)$ encodes how the mechanism locally resolves the remaining escape energy at $u$ when the current observation is anchored at $z$.
In particular, the family $z\mapsto\psi_z$ allows the local geometry used by the mechanism to vary with the state of the escape process itself.

\begin{assumption}[Observable field regularity]
\label{ass:field_regularities_mechanism}
For every $z\in\mathbb R^n$, the observable field $\psi_z$ is locally Lipschitz on $U_z$.
Moreover, on the noncritical set $\{z\in\mathbb R^n:s_{\psi_z}(z)>0\}$, the maps
$z\mapsto w_{\psi_z}(z)$ and $z\mapsto s_{\psi_z}(z)$
are Borel measurable.
\end{assumption}

Assumption~\ref{ass:field_regularities_mechanism} imposes the basic regularity required for the observable geometry to be well defined on the ambient space.
Local Lipschitz regularity gives each anchored field a coherent first-order nonsmooth structure near its base point, while measurability ensures controlled variation with the anchor.
Hence the observable geometry can be treated as a stable ambient state-dependent object along realized escapes.

A mesoscopic organization of the observable geometry is introduced by fixing a directional resolution $\theta\in(0,\pi)$ and a multiplicative steepness resolution $\varrho>1$.
Let
\[
V_\theta=\{v_1,\dots,v_{N(\theta)}\}\subset\mathbb S^{n-1}
\]
be a finite maximal $\theta$-separated set.
By maximality, $V_\theta$ is automatically a finite $\theta$-net on $\mathbb S^{n-1}$.
Let
$Q_\theta:\mathbb S^{n-1}\to V_\theta$
be the nearest-point quantizer with a fixed tie-break rule.
For each ambient point $z$, apply the Clarke steepness--direction construction to the scalar field $\psi_z$ at its base point $z$.
Thus, on the noncritical set $\{z:s_{\psi_z}(z)>0\}$, one has the direction--steepness readout $w_{\psi_z}(z)$ and $s_{\psi_z}(z)$.
The corresponding finite-resolution direction label is
\[
v_{\psi_z}(z):=Q_\theta(w_{\psi_z}(z)),
\]
and the steepness label is the unique integer $k_{\psi_z}(z)\in\mathbb Z$ such that
\[
\varrho^{k_{\psi_z}(z)}\le s_{\psi_z}(z)<\varrho^{k_{\psi_z}(z)+1}.
\]
The steepness scale is quantized multiplicatively because steepness is a positive scale quantity whose binwise variation is controlled by the fixed factor $\varrho$ while every noncritical bin retains a positive lower bound.
By Assumption~\ref{ass:field_regularities_mechanism}, both maps
$z\mapsto v_{\psi_z}(z)$ and $z\mapsto k_{\psi_z}(z)$ are Borel measurable on $\{z:s_{\psi_z}(z)>0\}$.
They therefore induce a mesoscopic ambient partition by almost-everywhere label stability.

\begin{definition}[Observable cell]
\label{def:observable_cell}
An observable cell is a measurable set $C\subset\mathbb R^n$ with $\mathcal L^n(C)>0$ such that:
\begin{enumerate}
\item[\textnormal{(i)}] there exist labels $v\in V_\theta$ and $k\in\mathbb Z$ such that, for $\mathcal L^n$-a.e.\ $z\in C$, one has
\[
s_{\psi_z}(z)>0,\qquad v_{\psi_z}(z)=v,\qquad k_{\psi_z}(z)=k;
\]

\item[\textnormal{(ii)}] $C$ is $\mathcal L^n$-essentially connected;

\item[\textnormal{(iii)}] $C$ is $\subseteq_{\mathcal L^n}$-maximal among measurable sets with properties \textnormal{(i)}--\textnormal{(ii)}.
\end{enumerate}
\end{definition}

An observable cell is the maximal essentially connected ambient region on which the observable geometry has the same finite-resolution directional regime and the same steepness scale almost everywhere.
It is therefore the basic mesoscopic unit of observable geometric coherence. Within such a region, the observed escape geometry undergoes no reorganization at the chosen resolution, up to a null set.
Write the family of observable cells as $\{C_\bbb^\psi\}_{\bbb\in \mathfrak I^\psi}$, and for each $\bbb\in \mathfrak I^\psi$ write $(v_\bbb^\psi,k_\bbb^\psi)$ for the corresponding labels.

The observable cell decomposition fixes the mesoscopic finite-resolution geometry relative to which realized escape is described.
In the deterministic baseline considered here, it also fixes the local directional regime within which realized escape is carried out.
For a realized admissible path $\xi\in\mathcal A_x$ and a time $t$ such that $\dot\xi(t)\neq 0$, write
\[
w_\xi(t):=\frac{\dot\xi(t)}{\|\dot\xi(t)\|}
\]
for the realized unit direction of $\xi$ at $t$.
The basic requirement imposed below is that, whenever the realized path passes through an observable cell $C_\bbb^\psi$, its local direction remain within the directional regime encoded by $v_\bbb^\psi$, and the anchored observable field be strictly descending along the realized increment.
Thus the observable partition provides both the mesoscopic readout of observable escape geometry and the finite-resolution regime for local escape realization.

\begin{assumption}[Observable-consistent realization]
\label{ass:observable_consistent_realization}
For every $x\in\mathcal M$ and every realized admissible path $\xi\in\mathcal A_x$, for a.e.\ $t$ in
the parameter domain of $\xi$ with $\dot\xi(t)\neq0$, if $\xi(t)\in C_\bbb^\psi$ for some
$\bbb\in \mathfrak I^\psi$, then
\[
Q_\theta\!\bigl(w_\xi(t)\bigr)=v_\bbb^\psi,
\qquad
\psi_{\xi(t)}^\circ(\xi(t);\dot\xi(t))<0.
\]
\end{assumption}

Assumption~\ref{ass:observable_consistent_realization} requires realized escape to be carried out under the observable geometry itself.
The quantized direction condition states that the realized increment lies in the finite-resolution directional regime encoded by the current observable cell.
The negativity condition states that, within this regime, the anchored observable field is strictly descending along the realized increment.

The main line is formulated under two deliberate reductions.
It retains only the deterministic geometric skeleton of realized escape and takes the current ambient point as the primitive state variable.
These choices isolate the geometric content relevant to the subsequent theory while suppressing additional realization-side and state-space complexity.
The following remarks record these suppressed layers.

\begin{remark}[Stochastic realization]
\label{rem:realization_side_refinements}
The present formulation isolates the deterministic geometric core of realized escape.
Randomness at the realization level may be incorporated by assigning to each ambient point $z$ a Borel probability kernel $\Pi_z$ describing the selection of local increments.
If, conditionally at each realized state, the cell-consistent descending regime in Assumption~\ref{ass:observable_consistent_realization} is selected with probability at least $q\in(0,1]$, then the deterministic realized-dynamics conclusions acquire a probabilistic reading.
Indeed, for any decomposition of a realized path into finitely many coupled-cell or frontier pieces, the event that all pieces follow the prescribed regime is bounded below by a corresponding power of $q$.
Thus stochasticity does not change the geometric content of the theory; it only weights the deterministic conclusions by realization probability.
\end{remark}

\begin{remark}[History dependence]
\label{rem:state_augmentation_mechanism}
The present formulation takes the current ambient point as the primitive state variable.
This isolates the geometric core: the local observable field attached to the current point, the realized ambient increment, and the ambient extension $\hat\phi$ of the escape geometry.
If history dependence is to be represented explicitly, one may augment the state to
$\mathbf z=(z,h)\in\mathcal Z$ with projection $\pi(\mathbf z)=z$.
The preceding constructions then lift through the ambient component.
Thus history dependence can be incorporated by state augmentation without changing the ambient geometric structure used here.
\end{remark}

\subsection{Coupled-Cell Delivery}
\label{subsec:coupled_cell_delivery}

Realized delivery is governed by the interaction between two finite-resolution geometries.
The observable cells describe the local regime under which the realized motion is generated, while the ambient cells describe how the same motion is measured by the ambient extension $\hat\phi$.
Their intersections form coupled cells, on which realized length admits a fixed conversion into decrease of $\hat\phi$.
Apply the same finite direction--steepness discretization and the same maximal essentially connected cell construction to the ambient extension $\hat\phi$.
This produces a family of ambient cells
$\{C_\aaa^{\hat\phi}\}_{\aaa\in \mathfrak I^{\hat\phi}}$,
and for each $\aaa\in \mathfrak I^{\hat\phi}$ the associated finite-resolution labels
$(v_\aaa^{\hat\phi},k_\aaa^{\hat\phi})$.

\begin{assumption}[Cellwise transverse Clarke regularity]
\label{ass:cellwise_transverse_clarke_regularity}
For every ambient cell $C_\aaa^{\hat\phi}$, there exists an $\mathcal L^n$-null set
$N_\aaa\subset C_\aaa^{\hat\phi}$ such that $\hat\phi$ has vanishing Clarke transverse spread at every
$z\in C_\aaa^{\hat\phi}\setminus N_\aaa$ with $s_{\hat\phi}(z)>0$.
Moreover, for every realized admissible path $\xi\in\mathcal A_x$, the set of times at which
$\xi(t)$ belongs to one of these exceptional sets has one-dimensional Lebesgue measure zero.
\end{assumption}

Assumption~\ref{ass:cellwise_transverse_clarke_regularity} makes the ambient readout first-order single-axis inside each ambient cell.
It rules out positive-measure transverse switching of the Clarke subdifferential of $\hat\phi$ along realized paths.
Consequently, the ambient direction label represents a stable descent axis rather than a merely selected element of a multi-directional nonsmooth geometry.

Fix $x\in\mathcal M$ and let $\xi\in\mathcal A_x$ be a realized admissible path.
For $t$ in the parameter domain of $\xi$, define the observable and ambient cell readouts by
\[
\bbb_\xi(t)=\bbb \iff \xi(t)\in C_\bbb^\psi,
\qquad
\aaa_\xi(t)=\aaa \iff \xi(t)\in C_\aaa^{\hat\phi}.
\]
The observable readout $\bbb_\xi$ records the local observable regime under which the realized motion is generated, whereas the ambient readout $\aaa_\xi$ records the location of the same path within the ambient extension of the escape geometry.
Taken together, these two pathwise cell assignments constitute the coupled readout.

\begin{definition}[Coupled cell]
\label{def:coupled_cell}
For $\aaa\in \mathfrak I^{\hat\phi}$ and $\bbb\in \mathfrak I^\psi$, define the associated coupled cell by
$
C_{\aaa\bbb}:=C_\aaa^{\hat\phi}\cap C_\bbb^\psi\subset\mathbb R^n$.
Write
$\mathfrak J
:=
\{(\aaa,\bbb)\in \mathfrak I^{\hat\phi}\times \mathfrak I^\psi:\ \mathcal L^n(C_{\aaa\bbb})>0\}$
for the coupled-cell catalog.
\end{definition}

\begin{assumption}[Cell-interface invisibility]
\label{ass:cell_interface_invisibility}
For every $x\in\mathcal M$ and every realized admissible path $\xi\in\mathcal A_x$, the set of times at which $\xi(t)$ lies on an ambient-cell interface or an observable-cell interface has one-dimensional Lebesgue measure zero.
\end{assumption}

A coupled cell is the basic ambient region on which a realized escape is read simultaneously through the observable geometry and through the ambient extension.
The coupled readout therefore resolves how the realized motion is generated locally and how that same motion advances relative to $\hat\phi$.
Assumption~\ref{ass:cell_interface_invisibility} places the realized path, up to a time-null set, on well-defined coupled cells, so that its progress may be analyzed on successive coupled-cell segments.

\begin{theorem}[Cell delivery law]
\label{thm:cell_delivery}
Let $x\in\mathcal M$ and let $\xi\in\mathcal A_x$ be a realized admissible path.
Let $I=[t_0,t_1]$ be a compact interval in the parameter domain of $\xi$ such that, for some
$(\aaa,\bbb)\in\mathfrak J$, one has
$\xi(t)\in C_{\aaa\bbb}$
for a.e.\ $t\in I$.
Set
\[
\Theta_{\aaa\bbb}
:=
\min\left\{
\pi,\,
d_{\angle}(v_\aaa^{\hat\phi},v_\bbb^\psi)+2\theta
\right\},
\qquad
\mathscr R_{\aaa\bbb}
:=
\varrho^{k_\aaa^{\hat\phi}}
\cos\Theta_{\aaa\bbb}.
\]
If $\Theta_{\aaa\bbb}\le \pi/2$,
then
\[
\hat\phi(\xi(t_0))-\hat\phi(\xi(t_1))
\ge
\mathscr R_{\aaa\bbb}
\int_I \|\dot\xi(t)\|\,dt .
\]
\end{theorem}

\begin{proof}
Let $N_\aaa$ be the exceptional set associated with the ambient cell
$C_\aaa^{\hat\phi}$ in
Assumption~\ref{ass:cellwise_transverse_clarke_regularity}.
By the pathwise time-null exceptional-set condition in
Assumption~\ref{ass:cellwise_transverse_clarke_regularity}, the set of times
$t\in I$ for which $\xi(t)\in N_\aaa$ has one-dimensional Lebesgue measure zero.

Fix $t$ in a full-measure subset of $I$ such that
$\xi(t)\in C_{\aaa\bbb}$, $\xi(t)\notin N_\aaa$, and either $\dot\xi(t)=0$ or
Assumption~\ref{ass:observable_consistent_realization} applies.
Then the ambient labels are defined, $s_{\hat\phi}(\xi(t))>0$, and
$\hat\phi$ has vanishing Clarke transverse spread at $\xi(t)$.

If $\dot\xi(t)=0$, positive homogeneity gives
\[
-\hat\phi^\circ(\xi(t);\dot\xi(t))=0,
\]
and the desired pointwise estimate is immediate.
Assume $\dot\xi(t)\neq0$ and write
\[
w_\xi(t):=\frac{\dot\xi(t)}{\|\dot\xi(t)\|}.
\]
Since $\xi(t)\in C_{\aaa\bbb}$, the ambient cell label gives
\[
d_{\angle}\!\bigl(w_{\hat\phi}(\xi(t)),v_\aaa^{\hat\phi}\bigr)\le \theta,
\qquad
\varrho^{k_\aaa^{\hat\phi}}\le s_{\hat\phi}(\xi(t))<\varrho^{k_\aaa^{\hat\phi}+1}.
\]
Observable consistency gives
\[
Q_\theta(w_\xi(t))=v_\bbb^\psi,
\]
and therefore
\[
d_{\angle}(w_\xi(t),v_\bbb^\psi)\le\theta .
\]
Since $\Theta_{\aaa\bbb}\le\pi/2$, the truncation in the definition of
$\Theta_{\aaa\bbb}$ is inactive. Hence
\[
d_{\angle}\!\bigl(w_{\hat\phi}(\xi(t)),w_\xi(t)\bigr)
\le
d_{\angle}(v_\aaa^{\hat\phi},v_\bbb^\psi)+2\theta
=
\Theta_{\aaa\bbb}.
\]
Thus
\[
\cos d_{\angle}\!\bigl(w_{\hat\phi}(\xi(t)),w_\xi(t)\bigr)
\ge
\cos\Theta_{\aaa\bbb}\ge0 .
\]

By Lemma~\ref{lem:cone_descent_vanishing_clarke_spread},
\[
-\hat\phi^\circ(\xi(t);\dot\xi(t))
\ge
s_{\hat\phi}(\xi(t))
\cos d_{\angle}\!\bigl(w_{\hat\phi}(\xi(t)),w_\xi(t)\bigr)
\|\dot\xi(t)\|.
\]
Using the ambient steepness bin and the angular estimate, we get
\[
-\hat\phi^\circ(\xi(t);\dot\xi(t))
\ge
\varrho^{k_\aaa^{\hat\phi}}
\cos\Theta_{\aaa\bbb}\,
\|\dot\xi(t)\|
=
\mathscr R_{\aaa\bbb}\|\dot\xi(t)\|.
\]
Thus this pointwise estimate holds for a.e.\ $t\in I$.

Since $t\mapsto\hat\phi(\xi(t))$ is absolutely continuous and the Clarke chain inequality gives
\[
-\frac{d}{dt}\hat\phi(\xi(t))
\ge
-\hat\phi^\circ(\xi(t);\dot\xi(t))
\qquad\text{for a.e.\ }t\in I,
\]
integration over $I=[t_0,t_1]$ yields
\[
\hat\phi(\xi(t_0))-\hat\phi(\xi(t_1))
\ge
\mathscr R_{\aaa\bbb}
\int_{t_0}^{t_1}\|\dot\xi(t)\|\,dt .
\]
\end{proof}

Theorem~\ref{thm:cell_delivery} identifies the cell delivery rate at which realized motion is converted into decrease of the ambient escape field. On a coupled cell, the observable readout fixes the finite-resolution direction regime followed by the path, while the ambient readout fixes the descent direction and steepness scale of $\hat\phi$. Thus $\mathscr R_{\aaa\bbb}$ is a local conversion rate from realized length to ambient escape progress. Its angular term measures directional compatibility, and the factor $\varrho^{k_\aaa^{\hat\phi}}$ measures the ambient steepness available in the cell. No observable steepness factor appears because the statement is length-based rather than an accounting of observable descent. The condition $\Theta_{\aaa\bbb}\le\pi/2$ isolates the forward regime in which realized length has a nonnegative projection onto the ambient descent geometry. Hence a coupled cell provides a local conversion law from realized path length to guaranteed decrease of $\hat\phi$, and the resulting pathwise estimate is obtained by concatenating these cell delivery rates along the realized frontier.

The cell delivery rate converts realized length on a coupled-cell segment into decrease of the ambient extension.
Endpoint perturbation is instead measured by the gauge between the initial point and the boundary hit.
A pathwise comparison therefore requires two structures: a gauge control by ordered path segments, and a decomposition that separates the segments carrying the decrease of $\hat\phi$ from complementary motion along the same realization.

\begin{assumption}[Path-length gauge control]
\label{ass:path_length_gauge_control}
There exists a constant $c_p>0$ such that, for every absolutely continuous path $\zeta:[a,b]\to\mathbb R^n$,
\[
p\bigl(\zeta(a),\zeta(b)\bigr)
\le
c_p\int_a^b \|\dot\zeta(t)\|\,dt.
\]
\end{assumption}

Assumption~\ref{ass:path_length_gauge_control} places the endpoint perturbation gauge under pathwise control along realized motion.
It states that the gauge of a path segment is controlled by its ambient length, and thus gives $p$ a coherent pathwise meaning.
The frontier structure then identifies the ordered segments carrying the full decrease of $\hat\phi$ along the realized path.

\begin{definition}[Frontier decomposition]
\label{def:frontier_decomposition}
Let $x\in\mathcal M$ and let $\xi\in\mathcal A_x$ satisfy
$\tau_{\partial\Omega}(\xi)<\infty$.
A frontier decomposition of $\xi$ up to $\tau_{\partial\Omega}(\xi)$ is a finite nonempty family of pairwise disjoint compact intervals
$\{I_j=[s_j,t_j]\}_{j=1}^N$
in the parameter domain of $\xi$ such that, for each $j$, there exists
$(\aaa_j,\bbb_j)\in\mathfrak J$ with
\[
\xi(t)\in C_{\aaa_j\bbb_j}
\qquad\text{for a.e.\ }t\in I_j,
\]
and
\[
\sum_{j=1}^N
\bigl(\hat\phi(\xi(s_j))-\hat\phi(\xi(t_j))\bigr)
=
\hat\phi(x)-\hat\phi(\xi(\tau_{\partial\Omega}(\xi))).
\]
\end{definition}

A frontier decomposition selects the coupled-cell segments on which the full drop of the ambient extension along the path is carried.
It separates frontier-carrying motion, where decrease of $\hat\phi$ is delivered, from complementary motion that accompanies the same realization without contributing to the prescribed drop.
A pathwise perturbation estimate then depends on whether the gauge burden of the complementary motion remains controlled by that of the frontier-carrying segments.

\begin{assumption}[Frontier-complement control]
\label{ass:frontier_complement_control}
There exists a constant $c_{\mathrm{gap}}\ge 0$ such that the following holds.
Let $x\in\mathcal M$ and let $\xi\in\mathcal A_x$ satisfy
$\tau_{\partial\Omega}(\xi)<\infty$.
Let $\{I_j=[s_j,t_j]\}_{j=1}^N$ be a frontier decomposition of $\xi$ up to
$\tau_{\partial\Omega}(\xi)$.
Let $\{G_\ell\}_{\ell=1}^M$ be the connected components of
\[
[0,\tau_{\partial\Omega}(\xi)]\setminus \bigcup_{j=1}^N I_j .
\]
Then
\[
\sum_{\ell=1}^M
p\bigl(\xi(\inf G_\ell),\xi(\sup G_\ell)\bigr)
\le
c_{\mathrm{gap}}
\sum_{j=1}^N p\bigl(\xi(s_j),\xi(t_j)\bigr).
\]
\end{assumption}

Assumption~\ref{ass:frontier_complement_control} imposes pathwise gauge comparability between complementary motion and frontier-carrying motion.
The total perturbation gauge of the complementary motion is controlled by the gauge carried by the segments delivering decrease of $\hat\phi$, up to the uniform factor $c_{\mathrm{gap}}$.
Thus the gauge of the realized path is controlled by its frontier-carrying part.
Combined with the local coupled-cell delivery law, this yields a realization-scale estimate for the successful perturbation gauge.

\begin{corollary}[Reference-scale realization efficiency]
\label{cor:reference_scale_realization_efficiency}
Let $x\in\mathcal M$ and let $\xi\in\mathcal A_x$ satisfy
$\tau_{\partial\Omega}(\xi)<\infty$.
Let $\{I_j=[s_j,t_j]\}_{j=1}^N$ be a frontier decomposition of $\xi$ up to
$\tau_{\partial\Omega}(\xi)$, where
$\xi(t)\in C_{\aaa_j\bbb_j}$
for a.e.\ $t\in I_j$.
Set
$\mathscr R_\ast:=\min_{1\le j\le N}\mathscr R_{\aaa_j\bbb_j}$.
If $\mathscr R_\ast>0$, then
\[
p\bigl(x,\xi(\tau_{\partial\Omega}(\xi))\bigr)
\le
\frac{(1+c_{\mathrm{gap}})\,c_p}{\mathscr R_\ast}\,\phi(x).
\]
\end{corollary}

\begin{proof}
Write $\tau:=\tau_{\partial\Omega}(\xi)$ and
$\mathscr R_j:=\mathscr R_{\aaa_j\bbb_j}$.
Since $\mathscr R_\ast>0$, one has $\mathscr R_j>0$ for every $j$.
As $\mathscr R_j=\varrho^{k_{\aaa_j}^{\hat\phi}}\cos\Theta_{\aaa_j\bbb_j}$ with
$\Theta_{\aaa_j\bbb_j}\in[0,\pi]$, this implies $\Theta_{\aaa_j\bbb_j}<\pi/2$.
Hence Theorem~\ref{thm:cell_delivery} applies on each frontier interval $I_j$.

For $I_j=[s_j,t_j]$, Theorem~\ref{thm:cell_delivery} and
Assumption~\ref{ass:path_length_gauge_control} give
\[
p\bigl(\xi(s_j),\xi(t_j)\bigr)
\le
c_p\int_{s_j}^{t_j}\|\dot\xi(t)\|\,dt
\le
\frac{c_p}{\mathscr R_j}
\bigl(\hat\phi(\xi(s_j))-\hat\phi(\xi(t_j))\bigr).
\]
Since $\mathscr R_j\ge\mathscr R_\ast$, summing over $j$ yields
\[
\sum_{j=1}^N p\bigl(\xi(s_j),\xi(t_j)\bigr)
\le
\frac{c_p}{\mathscr R_\ast}
\sum_{j=1}^N
\bigl(\hat\phi(\xi(s_j))-\hat\phi(\xi(t_j))\bigr).
\]
By the frontier decomposition property,
\[
\sum_{j=1}^N
\bigl(\hat\phi(\xi(s_j))-\hat\phi(\xi(t_j))\bigr)
=
\hat\phi(x)-\hat\phi(\xi(\tau)).
\]
Using $\hat\phi(x)=\phi(x)$ and $\hat\phi\ge0$, we obtain
\[
\sum_{j=1}^N p\bigl(\xi(s_j),\xi(t_j)\bigr)
\le
\frac{c_p}{\mathscr R_\ast}\,\phi(x).
\tag{$*$}
\]

Let $\{G_\ell\}_{\ell=1}^M$ be the connected components of
$[0,\tau]\setminus\bigcup_{j=1}^N I_j$.
By finite subadditivity of $p$ along the ordered frontier and gap intervals, and by
Assumption~\ref{ass:frontier_complement_control},
\[
\begin{aligned}
p\bigl(x,\xi(\tau)\bigr)
&\le
\sum_{j=1}^N p\bigl(\xi(s_j),\xi(t_j)\bigr)
+
\sum_{\ell=1}^M
p\bigl(\xi(\inf G_\ell),\xi(\sup G_\ell)\bigr)  \\
&\le
(1+c_{\mathrm{gap}})
\sum_{j=1}^N p\bigl(\xi(s_j),\xi(t_j)\bigr).
\end{aligned}
\]
Combining this with ($*$) gives
\[
p\bigl(x,\xi(\tau)\bigr)
\le
\frac{(1+c_{\mathrm{gap}})\,c_p}{\mathscr R_\ast}\,\phi(x).
\]
\end{proof}

Corollary~\ref{cor:reference_scale_realization_efficiency} can be viewed as a necessary structural
condition for successful realized escape at the reference scale set by $\phi(x)$.
The quantity $\mathscr R_{\aaa\bbb}$ measures the decrease of $\hat\phi$ delivered per unit realized path length on a frontier coupled cell.
The estimate shows that endpoint perturbation remains commensurate with $\phi(x)$ only when this delivered decrease is not diluted along the realized path: the frontier-carrying cells must have sufficiently large cell delivery rate, and the complementary motion must not carry a disproportionate gauge burden.
In this sense, the factor $\mathscr R_\ast^{-1}$ records weak local delivery on the frontier, while $1+c_{\mathrm{gap}}$ records the path-level dilution of that delivery by non-frontier motion.
Thus, whenever $\phi(x)$ is small but the successful endpoint perturbation is large, the discrepancy is already encoded in the realized escape geometry of the path, and indicates a low effective delivery rate relative to the ambient extension.

\begin{remark}[Cost and backtracking]
\label{rem:cost_and_backtracking}
The cell delivery rate also gives a cost interpretation.
On a forward coupled-cell segment, where $\Theta_{\aaa\bbb}\le\pi/2$, the reciprocal
$\mathscr R_{\aaa\bbb}^{-1}$ measures the realized-length cost required to produce a fixed decrease of $\hat\phi$.
Low delivery efficiency therefore means that realized length is spent with limited ambient escape progress.
The complementary regime $\Theta_{\aaa\bbb}>\pi/2$ is excluded from the positive-delivery statement of Theorem~\ref{thm:cell_delivery}, but it is described by the same cell comparison in signed form.
There the realized direction has nonpositive projection onto the ambient descent geometry of $\hat\phi$, even though it may remain admissible, or locally descending, for the observable field.
Backtracking is therefore a geometric mismatch between realization and ambient readout: motion is generated under the observable geometry, while its delivered component relative to $\hat\phi$ is neutral or negative.
\end{remark}

\begin{remark}[Persistence of delivered decrease]
\label{rem:persistence_of_delivered_decrease}
The coupled-readout law also has a structural implication beyond a single ambient extension.
The relevant phenomenon is the stable readability of escape success across ambient readings, path classes, or nearby initial points.
As a base requirement, realized motion must be converted into decrease of the ambient extension under which the realization is read.
For shared escape success, the critical structure is the persistence of that delivered decrease when the same path, or a nearby realization in the same path class, is read through another ambient extension.
This distinction separates structural transfer from endpoint coincidence.
In a structurally stable setting, where success is not produced by an isolated agreement of terminal points, shared success is carried by persistence of delivered decrease along the path.
Thus the relevant object is not only the terminal boundary hit, but the survival of delivered decrease under a change of ambient readout.
\end{remark}

\subsection{Interfacial Branch Leakage}
\label{subsec:interfacial_branch_leakage}

Realized escape is not only a pointwise object.
Nearby initial points on $\mathcal M$ may follow realized paths that either preserve a coherent observable organization or separate into distinct observable modes.
Observable-cell histories provide the natural language for describing this transition from coherent escape to branching.

Fix a connected measurable region $P\subset\mathcal M$ with $\mu(P)>0$.
For $\mu$-a.e.\ $x\in P$, let
$\xi_x$ be a realized admissible escape segment parameterized by arc length.
Fix an arc-length scale $t_\ast>0$, and restrict each segment to $[0,t_\ast]$, so that
\[
\xi_x:[0,t_\ast]\to\mathbb R^n,
\qquad
\|\xi_x'(t)\|=1
\quad\text{for a.e.\ }t\in[0,t_\ast].
\]
For a measurable set $A\subset P$ and $t\in[0,t_\ast]$, write
\[
\ZZ_t(A):=\{\xi_x(t):x\in A\}
\]
for the realized trace of $A$ at arc length $t$.
For $t\in[0,t_\ast]$ and $x,x'\in P$, define
\[
x\sim_t x'
\iff
\bbb_{\xi_x}(r)=\bbb_{\xi_{x'}}(r)
\quad\text{for a.e.\ }r\in[0,t].
\]
For each $t\in[0,t_\ast]$, the equivalence relation $\sim_t$ partitions $P$ into observable escape branches at arc length $t$.
% Since the observable-cell catalog is countable modulo $\mathcal L^n$-null sets, this branch family is countable modulo $\mu$-null sets.

\begin{definition}[Dominant branch mass]
\label{def:dominant_branch_mass}
For $t\in[0,t_\ast]$, let
\[
\mathcal K_t(P):=P/{\sim_t}
\]
be the partition of $P$ into observable escape branches at arc length $t$.
The dominant branch mass of $P$ at arc length $t$ is
\[
D_P(t):=\sup_{B\in\mathcal K_t(P)}\mu(B).
\]
\end{definition}

The quantity $D_P(t)$ is measured on the initial region $P$.
It is the largest $\mu$-mass of initial points whose realized paths share one observable-cell history on $[0,t]$.
Because branches are defined by accumulated observable history, the partitions $\mathcal K_t(P)$ refine as $t$ increases.
A decrease of $D_P$ is therefore a loss of common observable history within the initial region.
Such loss occurs when realized paths separate into different observable-cell sequences, and its geometric carrier is the crossing of observable-cell interfaces.

For observable cells $C_\bbb^\psi$ and $C_{\bbb'}^\psi$, 
we say that $\bbb'$ is adjacent to $\bbb$, and write
$\bbb'\in\adj(\bbb)$, if there exists a measurable set
\[
\mathit{\Pi}_{\bbb\bbb'}
\subset
\partial C_\bbb^\psi\cap\partial C_{\bbb'}^\psi
\]
with $\mathcal H^{n-1}(\mathit{\Pi}_{\bbb\bbb'})>0$
that is covered, up to an $\mathcal H^{n-1}$-null set, by countably many locally Lipschitz hypersurface pieces and locally separates
$C_\bbb^\psi$ from $C_{\bbb'}^\psi$.
Such a set is a regular crossing interface along which realized escape may pass from one observable cell to another.

For an adjacent pair $(\bbb,\bbb')$, let
$n_{\bbb\to\bbb'}(z)$ denote a measurable unit normal field on
$\mathit{\Pi}_{\bbb\bbb'}$, defined for $\mathcal H^{n-1}$-a.e.\ $z$, oriented from
$C_\bbb^\psi$ toward $C_{\bbb'}^\psi$.
For $x\in P$, an adjacent pair $(\bbb,\bbb')$ and an interval
$0\le t_0<t_1\le t_\ast$, define the interval first-crossing arc length
\[
\sigma_{\bbb\to\bbb'}^{t_0,t_1}(x)
:=
\inf\Bigl\{
t\in[t_0,t_1]:
\xi_x(t)\in \mathit{\Pi}_{\bbb\bbb'},
\ \xi_x \text{ crosses from } C_\bbb^\psi
\text{ to } C_{\bbb'}^\psi
\text{ through } \mathit{\Pi}_{\bbb\bbb'}
\text{ at } t
\Bigr\},
\]
with the convention $\sigma_{\bbb\to\bbb'}^{t_0,t_1}(x)=+\infty$ if no such crossing occurs.
Thus $\sigma_{\bbb\to\bbb'}^{t_0,t_1}$ records the first regular crossing from
$C_\bbb^\psi$ to $C_{\bbb'}^\psi$ through the chosen interface
$\mathit{\Pi}_{\bbb\bbb'}$ during the arc-length interval $[t_0,t_1]$.

\begin{assumption}[Regionwise branch regularity]
\label{ass:regionwise_branch_regularities}
Fix $t_\ast>0$.
The realized escape family over $P$ is regular on $[0,t_\ast]$ in the following sense.

\begin{enumerate}
\item[\textnormal{(i)}] \textnormal{Observable-history structure.}
For every $t\in[0,t_\ast]$, the observable-history partition
$\mathcal K_t(P)$ is countable modulo $\mu$-null sets, and the map
$(x,t)\mapsto \xi_x(t)$ is measurable on $A\times[0,t_\ast]$ for every measurable
$A\subset P$.
Modulo a $\mu$-null subset of $P$, every change of observable-cell readout along
$\xi_x$ on $[0,t_\ast]$ occurs through a chosen regular crossing interface
$\mathit{\Pi}_{\bbb\bbb'}$ for some adjacent pair $(\bbb,\bbb')$.

\item[\textnormal{(ii)}] \textnormal{Regular crossings.}
For every measurable $A\subset P$, every adjacent pair $(\bbb,\bbb')$, and every
$0\le t_0<t_1\le t_\ast$, the interval crossing map
\[
x\longmapsto
\bigl(
\sigma_{\bbb\to\bbb'}^{t_0,t_1}(x),\,
\xi_x(\sigma_{\bbb\to\bbb'}^{t_0,t_1}(x))
\bigr)
\]
is measurable on
$A\cap\{\sigma_{\bbb\to\bbb'}^{t_0,t_1}<\infty\}$
and has uniformly bounded multiplicity on its image.
The interface $\mathit{\Pi}_{\bbb\bbb'}$ admits a measurable unit normal field
$n_{\bbb\to\bbb'}$ defined for $\mathcal H^{n-1}$-a.e.\ point and oriented from
$C_\bbb^\psi$ toward $C_{\bbb'}^\psi$.
Crossings for which the source-side incoming unit direction is undefined or fails to belong to the direction class labeled by $v^\psi_\bbb$ are excluded modulo $\mu$.

\item[\textnormal{(iii)}] \textnormal{Trace regularity.}
For every measurable $A\subset P$ and for a.e.\ $t\in[0,t_\ast]$, the trace
$\ZZ_t(A)$ is countably $m$-rectifiable, and the trace law
\[
\nu_t^A:=(x\mapsto \xi_x(t))_\#(\mu\!\lfloor_A)
\]
is absolutely continuous with respect to
$\mathcal H^m\!\lfloor_{\ZZ_t(A)}$, with uniformly bounded Radon--Nikodym density.

\item[\textnormal{(iv)}] \textnormal{Normal-trace regularity of crossings.}
For every measurable $A\subset P$, every adjacent pair $(\bbb,\bbb')$, and every
$0\le t_0<t_1\le t_\ast$, the first source-to-target crossings from
$C^\psi_\bbb$ to $C^\psi_{\bbb'}$ define a crossing measure on the spacetime
interface $[t_0,t_1]\times\mathit{\Pi}_{\bbb\bbb'}$. This crossing measure is
absolutely continuous with respect to the spacetime interfacial exposure measure
of $\ZZ_t(A)$, and its density is bounded, up to a uniform constant
$c_{\mathrm{flux}}$, by the positive source-side normal crossing speed.
\end{enumerate}
\end{assumption}

Assumption~\ref{ass:regionwise_branch_regularities} places the realized family over the initial region in a nondegenerate geometric regime.
It ensures that observable branching can be read from the geometry of realized traces.
Observable histories are measurable and countably organized, and their readout changes are exhausted, modulo $\mu$-null sets, by the chosen regular crossing interfaces.
The associated crossing maps are measurable with controlled multiplicity, and the interfaces carry oriented measurable normal fields.
At almost every realization time, the image of any subregion remains an $m$-dimensional trace with bounded pushed-forward density.
The normal-trace condition rules out singular crossing mass by requiring crossings across an observable interface to be carried by the spacetime interfacial exposure of the realized trace, with density controlled up to a uniform flux constant by the positive source-side normal crossing speed.
Under these conditions, loss of dominant branch mass can be estimated by outward flux across the chosen observable interfaces.

\begin{lemma}[Interface crossing leakage]
\label{lem:interface_crossing_leakage}
Let $A\subset P$ be measurable, let $0\le t_0<t_1\le t_\ast$, and let
$(\bbb,\bbb')$ be an adjacent pair.
Set
\[
F_{\bbb\to\bbb'}(A;t_0,t_1)
:=
A\cap\bigl\{x\in P:\ \sigma_{\bbb\to\bbb'}^{t_0,t_1}(x)<\infty\bigr\}.
\]
Then there exists a constant $c_{\mathrm{flux}}>0$ such that
\[
\mu\bigl(F_{\bbb\to\bbb'}(A;t_0,t_1)\bigr)
\le
c_{\mathrm{flux}}
\int_{t_0}^{t_1}
\int_{\ZZ_t(A)\cap\mathit{\Pi}_{\bbb\bbb'}}
\sup_{\substack{w\in\mathbb S^{n-1}\\ Q_\theta(w)=v_\bbb^\psi}}
\bigl(w\cdot n_{\bbb\to\bbb'}(z)\bigr)_+\,
d\mathcal H^{m-1}(z)\,dt.
\]
\end{lemma}

\begin{proof}
Write $F:=F_{\bbb\to\bbb'}(A;t_0,t_1)$ and define
\[
\rho_{\bbb\to\bbb'}(z)
:=
\sup_{\substack{w\in\mathbb S^{n-1}\\ Q_\theta(w)=v_\bbb^\psi}}
\bigl(w\cdot n_{\bbb\to\bbb'}(z)\bigr)_+,
\qquad z\in\mathit{\Pi}_{\bbb\bbb'} .
\]
For $x\in F$, set
\[
e_{\bbb\to\bbb'}^{t_0,t_1}(x)
:=
\bigl(
\sigma_{\bbb\to\bbb'}^{t_0,t_1}(x),\,
\xi_x(\sigma_{\bbb\to\bbb'}^{t_0,t_1}(x))
\bigr).
\]
By definition,
\[
e_{\bbb\to\bbb'}^{t_0,t_1}(F)
\subset
[t_0,t_1]\times\mathit{\Pi}_{\bbb\bbb'} .
\]
Assumption~\ref{ass:regionwise_branch_regularities}~(ii) gives measurability of this crossing map and
uniformly bounded multiplicity on its image.

At regular crossing points, the oriented normal $n_{\bbb\to\bbb'}$ is defined, and the crossing is from
$C_\bbb^\psi$ toward $C_{\bbb'}^\psi$.
Endpoint crossings and nonregular crossings are ignored modulo $\mu$ by
Assumption~\ref{ass:regionwise_branch_regularities}~(ii).
Since the paths are parameterized by arc length, the source-side positive normal crossing speed at a crossing time
$\sigma=\sigma_{\bbb\to\bbb'}^{t_0,t_1}(x)$ is
\[
\bigl(w_{\xi_x}^{-}(\sigma)\cdot n_{\bbb\to\bbb'}(\xi_x(\sigma))\bigr)_+ ,
\]
where $w_{\xi_x}^{-}(\sigma)$ denotes the source-side incoming unit direction.
By Assumption~\ref{ass:observable_consistent_realization}, together with the regular-crossing exclusion in
Assumption~\ref{ass:regionwise_branch_regularities}~(ii), the source-side incoming realized direction belongs to
the observable direction class of $C_\bbb^\psi$ at crossing times, modulo a $\mu$-null set.
Hence the source-side positive normal crossing speed is bounded by
\[
\rho_{\bbb\to\bbb'}(\xi_x(\sigma)).
\]

Consider the spacetime trace
\[
\Phi_A:A\times[t_0,t_1]\to[t_0,t_1]\times\mathbb R^n,
\qquad
\Phi_A(x,t):=(t,\xi_x(t)).
\]
For a.e.\ $t$, the pushed-forward trace law
\[
\nu_t^A=(x\mapsto\xi_x(t))_\#(\mu\!\lfloor_A)
\]
has uniformly bounded density with respect to
$\mathcal H^m\!\lfloor_{\ZZ_t(A)}$.
Thus initial $\mu$-mass transported by the family is represented on the realized
$m$-dimensional trace without singular concentration relative to the trace measure.
Assumption~\ref{ass:regionwise_branch_regularities}~(iv) supplies the corresponding
normal-trace regularity for interface crossings: the crossing measure induced by
$e_{\bbb\to\bbb'}^{t_0,t_1}$ is absolutely continuous with respect to the spacetime
interfacial exposure measure of $\ZZ_t(A)$ on $\mathit{\Pi}_{\bbb\bbb'}$, and its density is controlled,
up to the uniform constant $c_{\mathrm{flux}}$, by the source-side positive normal crossing speed.
Since this speed is bounded by $\rho_{\bbb\to\bbb'}$ at regular crossings, the normal-trace bound gives
\[
\mu(F)
\le
c_{\mathrm{flux}}
\int_{t_0}^{t_1}
\int_{\ZZ_t(A)\cap\mathit{\Pi}_{\bbb\bbb'}}
\rho_{\bbb\to\bbb'}(z)\,
d\mathcal H^{m-1}(z)\,dt .
\]
Substituting the definition of $\rho_{\bbb\to\bbb'}$ yields
\[
\mu\bigl(F_{\bbb\to\bbb'}(A;t_0,t_1)\bigr)
\le
c_{\mathrm{flux}}
\int_{t_0}^{t_1}
\int_{\ZZ_t(A)\cap\mathit{\Pi}_{\bbb\bbb'}}
\sup_{\substack{w\in\mathbb S^{n-1}\\ Q_\theta(w)=v_\bbb^\psi}}
\bigl(w\cdot n_{\bbb\to\bbb'}(z)\bigr)_+\,
d\mathcal H^{m-1}(z)\,dt .
\]
This proves the claim.
\end{proof}

Lemma~\ref{lem:interface_crossing_leakage} provides the local control of branch loss through a fixed observable-cell interface.
For the dominant branch profile, the relevant quantity is the cumulative outward leakage contributed by all interfaces adjacent to a given observable cell.
It is therefore convenient to collect these contributions at each boundary point into a single quantity.
For $z\in\partial C_\bbb^\psi$, define the total outward leakage strength by
\[
\mathscr L_\bbb(z)
:=
\sum_{\bbb'\in \adj(\bbb)}
\mathbf 1_{\mathit{\Pi}_{\bbb\bbb'}}(z)\,
\sup_{\substack{w\in\mathbb S^{n-1}\\ Q_\theta(w)=v_\bbb^\psi}}
\bigl(w\cdot n_{\bbb\to\bbb'}(z)\bigr)_+.
\]
Integrated against the realized trace of the family, it gives the geometric quantity that governs the decay of the dominant branch mass.

\begin{theorem}[Dominant-branch leakage law]
\label{thm:dominant_branch_leakage_law}
There exists a constant $c_{\mathrm{reg}}>0$, depending only on the regularity constants in
Assumption~\ref{ass:regionwise_branch_regularities} and the flux constant in
Lemma~\ref{lem:interface_crossing_leakage}, such that
\[
-\frac{d}{dt}D_P(t)
\le
c_{\mathrm{reg}}\,
\sum_{\bbb\in \mathfrak I^\psi}
\int_{\ZZ_t(P)\cap\partial C_\bbb^\psi}
\mathscr L_\bbb(z)\,d\mathcal H^{m-1}(z)
\qquad\text{in }\mathcal D'(0,t_\ast).
\]
\end{theorem}

\begin{proof}
Set
\[
\mathcal R_P(t)
:=
\sum_{\bbb\in \mathfrak I^\psi}
\int_{\ZZ_t(P)\cap\partial C_\bbb^\psi}
\mathscr L_\bbb(z)\,d\mathcal H^{m-1}(z).
\]
Fix $0\le t_0<t_1\le t_\ast$ and $\varepsilon>0$.
Choose $B_0\in\mathcal K_{t_0}(P)$ such that
\[
\mu(B_0)\ge D_P(t_0)-\varepsilon .
\]
Let $E\subset B_0$ be the set of points whose observable-cell readout changes on $[t_0,t_1]$.
Since $B_0$ is a single branch at time $t_0$, the points in $B_0\setminus E$ retain the same observable
history on $[0,t_1]$ and hence belong to a single branch of $\mathcal K_{t_1}(P)$.
Therefore
\[
D_P(t_1)\ge \mu(B_0\setminus E),
\]
and consequently
\[
D_P(t_0)-D_P(t_1)
\le
\mu(E)+\varepsilon .
\]

By the regular-interface completeness in
Assumption~\ref{ass:regionwise_branch_regularities}, every readout change in $E$ occurs, modulo a
$\mu$-null set, through one of the chosen regular crossing interfaces.
Thus
\[
E\subseteq_\mu
\bigcup_{\bbb\in \mathfrak I^\psi}
\bigcup_{\bbb'\in\adj(\bbb)}
F_{\bbb\to\bbb'}(B_0;t_0,t_1).
\]
Using subadditivity and Lemma~\ref{lem:interface_crossing_leakage}, and summing the adjacent-interface
contributions into $\mathscr L_\bbb$, we obtain
\[
\mu(E)
\le
c_{\mathrm{reg}}
\int_{t_0}^{t_1}
\sum_{\bbb\in\mathfrak I^\psi}
\int_{\ZZ_t(B_0)\cap\partial C_\bbb^\psi}
\mathscr L_\bbb(z)\,d\mathcal H^{m-1}(z)\,dt .
\]
Since $\ZZ_t(B_0)\subset \ZZ_t(P)$ for every $t$, this gives
\[
D_P(t_0)-D_P(t_1)
\le
c_{\mathrm{reg}}
\int_{t_0}^{t_1}\mathcal R_P(t)\,dt
+\varepsilon .
\]
Letting $\varepsilon\downarrow0$ yields the interval inequality
\[
D_P(t_0)-D_P(t_1)
\le
c_{\mathrm{reg}}
\int_{t_0}^{t_1}\mathcal R_P(t)\,dt .
\]

The partitions $\mathcal K_t(P)$ refine as $t$ increases, hence $D_P$ is nonincreasing and admits a
bounded-variation representative.
By the distributional convention in Section~\ref{subsec:prelim_integral_distributional}, the preceding
interval inequality is equivalent to
\[
-\frac{d}{dt}D_P(t)
\le
c_{\mathrm{reg}}\,\mathcal R_P(t)
\qquad\text{in }\mathcal D'(0,t_\ast).
\]
Substituting the definition of $\mathcal R_P(t)$ proves the theorem.
\end{proof}

Theorem~\ref{thm:dominant_branch_leakage_law} identifies the interface carrier of regionwise branching.
Although stated as a distributional differential inequality in the arc-length variable, it equivalently gives an interval-level control.
Over any subinterval of $[0,t_\ast]$, the loss of dominant-branch mass is bounded by the accumulated outward flux of the realized trace across the chosen observable interfaces.
Thus the decay of $D_P$ is controlled by geometric leakage across codimension-one interfaces of the observable partition.

The structure of the right-hand side is determined by two coupled factors.
The integration domain
$\ZZ_t(P)\cap\partial C_\bbb^\psi$
records the interfacial exposure of the realized trace.
The weight $\mathscr L_\bbb(z)$ records the outward component permitted by the observable direction at exposed interface points.
Dominant-branch mass can decay only where interfacial exposure and outward admissibility occur together.
Consequently, low branching does not mean that alternative observable modes are absent.
It means that the realized family has insufficient trace on separating interfaces, or insufficient outward admissibility on the exposed interface set.
The decaying quantity is therefore the $\mu$-mass of initial points that remain organized under one coherent observable mode of escape.

\begin{remark}[Branching and realization efficiency]
\label{rem:branching_realization_efficiency}
Observable branching should not be identified with inefficiency of individual realized escapes.
Under Assumption~\ref{ass:observable_consistent_realization}, realized motion is already constrained to descend within the current observable regime.
Loss of dominant-branch mass therefore does not arise from unconstrained path variation.
It reflects the relation between the observable partition and the escape geometry encoded by the ambient extension.

When the ambient extension has an effectively single dominant carrier of delivered decrease over the region, observable branching indicates that the observable partition has fragmented that carrier into several distinct histories.
In this case, branching records loss of readout fidelity, since one carrier of delivered decrease is represented by several observable branches.
When the ambient extension itself supports several competing carriers of delivered decrease over the region, branching has a different meaning.
It is then the observable trace of intrinsic multiplicity in the realized escape geometry, rather than a fragmentation introduced by the readout.

These two sources affect escape realization efficiency through the same structural mechanism.
Branching can weaken the possibility of assigning one coherent energy or perturbation scale to successful realization over the region.
In the readout-fracture case, a single delivered escape mechanism is split across several observable histories.
In the intrinsic-multiplicity case, nearby initial points are realized through genuinely different branches, whose delivery rates and perturbation burdens need not agree.
Thus branching does not necessarily indicate inefficient individual escape.
It indicates loss of uniform realization efficiency across the region.
As dominant-branch mass decays, delivered decrease becomes branch-dependent, and the energy or perturbation budget required for successful escape may vary across nearby initial points.
A fixed budget may therefore fail to induce stable escape behavior even inside a small initial region.
\end{remark}

\section{Learning-Theoretic Consequences}
\label{sec:ml_robustness_geometry}

The preceding construction is independent of a particular model class, attack algorithm, or
perturbation norm.
In learning systems, it organizes robustness through the relation between task retention, data
geometry, and admissible change.
The resulting viewpoint places robustness phenomena within a common measure-geometric escape structure.

\subsection{Instantiation of Geometric Primitives}
\label{subsec:ml_geometric_primitives}

The geometric objects developed in
Sections~\ref{sec:reference_escape_structure},
\ref{sec:shell_energy_axis}, and
\ref{sec:realized_escape_dynamics}
admit direct interpretations in standard learning settings.
They also specify regular regimes where score geometry, data geometry, perturbation budget, and
realization mechanism interact coherently.

In a learning problem, the scalar field $f$ represents a task-dependent retention score whose positive
region is the regime in which the target property is preserved.
For classification, this property is retention of the clean label.
With score vector $F=(F_1,\dots,F_K)$, one may localize the analysis to a clean class $y$ and take
$f(x):=F_y(x)-\max_{y'\neq y}F_{y'}(x)$,
or, when a specific competitor is relevant, the pairwise margin
$f(x):=F_y(x)-F_{y'}(x)$.
Then $\Omega=\{f>0\}$ is the region in which the target label, property, or constraint remains valid,
and $\partial\Omega=\{f=0\}$ is the retention boundary at which that validity is first lost.
The scalar field $f$ therefore does not impose a special model.
It is a scalarization of the task property whose loss is being studied.
In classification applications, one usually restricts $\mathcal M$ and $\mu$ to the cleanly retained
portion of the class under consideration, so that escape records the first loss of that retained label.

The reference manifold $\mathcal M$ represents the data-bearing geometry on which task retention,
vulnerable regions, and vulnerable mass are organized.
This interpretation is consistent with the manifold hypothesis, which takes high-dimensional data to
concentrate on or near a low-dimensional manifold and underlies much of manifold learning
~\citep{meilua2024manifold,tenenbaum2000global,belkin2003laplacian,narayanan2010sample}.
Depending on the application, $\mathcal M$ may represent a class-conditional data manifold, a semantic
manifold, a clean-data reference slice, or a lower-dimensional latent structure on which the relevant
data mass is concentrated~\citep{stutz2019disentangling}.
The measure $\mu$ is the corresponding reference measure on $\mathcal M$.
It need not coincide with the full ambient data distribution.
It may be a restricted law conditioned on a class, a domain, or another semantic attribute.
Thus the theory is not tied to a single global notion of data geometry.
It applies to the reference geometry on which robustness is to be evaluated.

The perturbation gauge $p$ records the notion of budget under which escape is measured.
In the classical setting, $p(z_1,z_2)=\|z_1-z_2\|_q$ for an ambient norm.
The formalism is not restricted to norm-induced budgets.
The same role may be played by a perceptual distance, a feature-space displacement, a latent editing effort, or a transport-type budget, provided that it satisfies the basic concatenation law.
Likewise, the admissible family $\mathcal A_x$ is not the set of all possible perturbation paths.
It is the chosen attack or intervention mechanism.
Projected first-order attacks, unconstrained penalty-based attacks, local-search or query-based attacks, patch or physical perturbation families, and editing trajectories in latent or generative models may all be encoded through the choice of $\mathcal A_x$~\citep{goodfellow2015explaining,madry2018towards,ilyas2018black,eykholt2018robust}.

Under this interpretation, the escape-energy field $\phi(x)$ is the minimal budget, under the chosen
admissible mechanism, required to bring the retained target property to its boundary from the reference
point $x\in\mathcal M$.
In classification, this is the minimal budget required to lose the retained clean label.
The escape set $X_\delta$ is the vulnerable set on the reference manifold at energy level $\delta$.
The retained property, perturbation gauge, and admissible mechanism jointly determine the effective escape geometry.
The shell-growth objects of Section~\ref{sec:shell_energy_axis} describe how vulnerability changes as
the budget is increased.
The shell $\mathit{\Sigma}_\delta$ consists of reference samples with minimal escape budget exactly
$\delta$, while the thin shell $\mathrm{Th}_{\delta,t}$ is the newly vulnerable region revealed when the
budget is raised from $\delta$ to $\delta+t$.

The component parameters $(d_\lambda,\beta_\lambda)$ encode the local geometric type of the effective
escape field on the reference data geometry.
The value $d_\lambda$ records the effective freedom, namely the dimension of locally energy-neutral
directions on the data manifold, while $\beta_\lambda$ records the order of transverse variation away from the shell base.
Thus $(d_\lambda,\beta_\lambda)$ describes how the retained property, the perturbation gauge, and the
admissible mechanism induce a local escape geometry over the active data geometry.
Several learning situations correspond to these local types.
\begin{enumerate}
\item[\textnormal{(i)}]
The case $d_\lambda=m-1$ and $\beta_\lambda=1$ is the regular margin-crossing regime.
It arises when the retained-score margin changes to first order in one effective transverse direction
and is neutral along the remaining shell directions.
This is the standard local picture for smooth models at noncritical margin points, and for
piecewise affine networks inside a fixed activation region when the active margin is nondegenerate on
the data manifold~\citep{montufar2014number,raghu2017expressive}.

\item[\textnormal{(ii)}]
The case $d_\lambda<m-1$ and $\beta_\lambda=1$ corresponds to a lower-dimensional active locus.
It may occur when vulnerability is concentrated on a semantic subpopulation, a latent slice, an
intersection of competitor margins, or a subset selected by mechanism constraints.
The escape field still varies linearly away from the active locus, but the locus is not
hypersurface-like inside the reference manifold.

\item[\textnormal{(iii)}]
The case $\beta_\lambda>1$ corresponds to higher-order contact between the effective escape geometry
and the reference geometry.
It may occur near tangential boundary contact, saturated or flattened score regions, symmetry or
invariance directions, critical latent-decoder directions, or smoothed surrogate objectives whose first
nonzero transverse variation is quadratic or higher order.
\end{enumerate}
These regimes are local configurations of a learned model and its chosen perturbation mechanism, not
separate global model classes, and may coexist within one model.

The assumptions on the reference geometry identify regular regimes in which shell generation is stable enough to be analyzed.
Assumption~\ref{ass:basic_regularity} is the baseline regularity requirement placed on the score
landscape, the perturbation budget, and the admissible mechanism.
For piecewise smooth or locally Lipschitz models, and for norm-like or transport-like perturbation
gauges, it is a natural regular regime rather than a model-specific restriction.
Assumption~\ref{ass:local_transverse_rise_escape_component} is more structural.
It does not assert that every point of every trained model carries one uniform local law.
It isolates shell patches on which the interaction between the effective escape geometry and the reference geometry has a coherent local order.
The resulting escape components are precisely the regions on which vulnerability is generated in one geometric mode.
Assumption~\ref{ass:asymptotically_stationary_local_rise} is stronger.
It is a rescaling hypothesis at active shell points.
In learning terms, after zooming to sufficiently small scales, the local boundary--gauge--mechanism interaction stabilizes to a fixed finite-scale profile along the transverse opening.
This is the regime in which a nondegenerate local morphology of shell growth can be represented.

The realized dynamics of Section~\ref{sec:realized_escape_dynamics} have a separate interpretation.
Once an attack path leaves the reference manifold, the manifold field $\phi$ no longer suffices to describe the remaining retention budget along the realized ambient trajectory.
The ambient extension $\hat\phi$ from Definition~\ref{def:ambient_extension} is therefore an analytical object that carries the reference escape geometry off the manifold.
It is not necessarily the objective used by the attack.
That role is played by the observable field assignment $z\mapsto \psi_z$, which represents the local surrogate geometry available to the mechanism at the current ambient point.
For gradient-based attacks, $\psi_z$ may be a local margin surrogate, or a sign-adjusted loss surrogate whose decrease corresponds to attack progress~\citep{goodfellow2015explaining,madry2018towards}.
For score-free or query-based attacks, it may represent a geometry reconstructed from local probing~\citep{ilyas2018black,uesato2018adversarial}.
For generative, latent, or editing-based attacks, it may encode a surrogate in the latent or decoded ambient space.
The main formulation treats this realization layer in a deterministic and memoryless baseline form, without identifying the ambient extension $\hat\phi$ with the observable mechanism $z\mapsto\psi_z$.

Assumptions~\ref{ass:ambient_extension_regularities} and \ref{ass:field_regularities_mechanism} express the existence of two distinct but coupled geometric fields: the ambient extension of the reference escape geometry and the local observable geometry used by the mechanism.
Assumption~\ref{ass:cellwise_transverse_clarke_regularity} is an analysis-level regular-regime condition on the ambient extension $\hat\phi$.
It ensures that, after finite-resolution organization, the ambient descent geometry is represented by stable first-order regimes along realized paths.
This condition is satisfied in smooth regimes and on the interiors of fixed-activation regions in standard piecewise smooth or piecewise affine networks.
It does not rule out nonsmooth learning models.
Rather, it excludes positive-measure switching regimes along realized paths where several active branches of $\hat\phi$ produce transverse first-order directions.

Assumption~\ref{ass:observable_consistent_realization} abstracts a broad family of first-order or locally guided attacks.
Its content is not that the mechanism knows the true ambient extension $\hat\phi$.
Its content is that the realized motion is generated by descent in the currently visible observable regime.
After the usual path interpolation of discrete iterates, this covers projected gradient methods,
penalty-based iterative attacks, steepest-ascent variants under a sign convention in which attack
progress is represented as decrease, and more general locally guided search procedures.
The stochasticity and state dependence suppressed in the main line of Section~\ref{sec:realized_escape_dynamics} are natural refinements of this deterministic skeleton.
Random restarts, noisy gradient estimates, stochastic proposal steps, or randomized local searches fall under the stochastic realization perspective of Remark~\ref{rem:realization_side_refinements}.
Momentum~\citep{dong2018boosting}, adaptive optimizers, beam or population search, and other history-dependent attack states fall under the augmented-state perspective of Remark~\ref{rem:state_augmentation_mechanism}.
Thus the deterministic ambient point formulation isolates the geometric core without excluding stateful or stochastic mechanisms.

The remaining assumptions organize pathwise efficiency and regionwise coherence.
Assumption~\ref{ass:path_length_gauge_control} requires that the endpoint perturbation budget of a realized attack segment is controlled by its ambient travel.
For norm-induced or comparable budgets, this is immediate.
Assumption~\ref{ass:frontier_complement_control} requires that successful realization is not dominated by unproductive ambient motion.
Its total budget burden may contain non-frontier motion, but that overhead remains controlled by the budget spent on the segments that deliver decrease of the ambient extension.
This is the nonpathological regime in which attack efficiency becomes geometrically readable.
Assumption~\ref{ass:regionwise_branch_regularities} expresses a mesoscopic regularity of attack families over neighboring points.
It ensures that loss of shared realized behavior can be read as a geometric transition across observable interfaces, rather than as an unstructured combinatorial fracture of nearby trajectories.
The normal-trace condition rules out singular branch loss by requiring crossings to be carried by the spacetime exposure of realized attack traces, with mass controlled by source-side normal crossing speed.

\subsection{Measure-Geometric Robustness}
\label{subsec:ml_escape_principle}

The preceding theory yields one robustness principle:
\begin{quote}
\emph{Robustness is the measure-geometric organization of escape induced by a retained task state, a reference measure, a perturbation gauge, and an admissible realization mechanism.}
\end{quote}
% robustness is the measure-geometric organization of escape induced by a retained task state, a
% reference measure, a perturbation gauge, and an admissible realization mechanism.
In a learning system, robustness is therefore not a scalar attribute of boundary proximity alone.
It concerns how possible escape is formed on the data-bearing geometry, how its generated mass and
finite-scale morphology are organized, and how a concrete mechanism realizes that escape through
ambient paths and observable readouts.

This principle has two inseparable parts.
The first is the reference organization of possible escape.
It describes how escape at a given energy scale is configured as a measurable structure on the
data-bearing geometry.
A small margin at one point is only a local signal; what matters is whether escape near a given budget
level remains exceptional or occupies a structured portion of the reference measure.
The shell structure records this layer by describing how vulnerability enters the reference measure as
the budget threshold varies, rather than merely where the boundary is first encountered.

The second part is the ambient realization of escape under a concrete mechanism.
An attack does not move in an abstract space of all possible perturbations.
It follows an observable surrogate, a gradient estimate, a query response, a latent editing direction,
or another mechanism-dependent geometry.
Thus the presence of budget-relevant escape in the reference geometry does not by itself determine how
that escape is accessed.
The realized layer records this mechanism-dependent access by separating the geometry followed by the
mechanism from the ambient escape geometry that measures delivered progress.

The role of the escape principle is to place these two parts in one measure-geometric framework.
Margin, boundary distance, vulnerable mass, verification refinement, attack efficiency, transferability,
and finite-scale morphology are different projections of the same escape structure.
They differ by whether one reads the formation of possible escape on the reference geometry, the finite-scale
morphology of generated vulnerability, the delivery of realized progress, or the persistence of escape
under changes of readout or mechanism.
The discussion below follows several projections where this measure-geometric content becomes
especially visible.

\emph{Vulnerable mass is generated along the robustness curve.}
For a retained task property and a reference measure $\mu$, the quantity $\mu(X_\delta)$ is the fraction of
reference samples that can lose the retained property within perturbation budget $\delta$ under the chosen
realization mechanism.
Equivalently, on a normalized reference measure, it is the failure-mass side of the budget-dependent
robustness curve.
The local object is therefore the incremental vulnerable mass revealed when the budget increases from $\delta$ to $\delta+t$.
This local budget response is not determined by nearest-boundary distance alone.
It depends on how the model score, data geometry, perturbation budget, and mechanism jointly generate
newly vulnerable samples near the current budget level.
Proposition~\ref{prop:fixed_base_generation} identifies the stable local source from which this newly
vulnerable mass is generated, and Theorem~\ref{thm:componentwise_increment_law} gives the componentwise
increment law
\[
\Delta_{\delta,\lambda}(t)\asymp t^{\gamma_\lambda},
\qquad
\gamma_\lambda=\frac{m-d_\lambda}{\beta_\lambda},
\qquad t\downarrow0 .
\]
Corollary~\ref{cor:interval_spectral_recovery} then shows that interval-level vulnerable mass is
recovered from the active growth orders along the budget axis.

This gives a measure-geometric reading of robustness evaluation.
Geometric analyses of robustness have already shown that vulnerability is tied to the interaction
between models, perturbation sets, and the geometry of the data region
~\citep{fawzi2018analysis,moosavi2017universal}.
The measure-geometric theory developed here refines this interaction into a local generation law for
the vulnerable fraction of the reference distribution.
The relevant object is the local budget response of the robustness curve.
This is consistent with recent critiques of evaluating robustness only through adversarial accuracy at
one prescribed radius~\citep{olivier2023how}.
Such an evaluation records only one point of the curve
$\delta\mapsto\mu(X_\delta)$, rather than the local law by which vulnerable mass changes with the
budget.
Adversarial training and min--max robust optimization~\citep{madry2018towards} can be
understood as acting on distribution-level robust risk, and hence as reshaping the budget-dependent
vulnerable-mass curve rather than only enlarging isolated pointwise margins.
Defense formulations based on robustness--accuracy tradeoffs and robust generalization support this
reading, since robustness gains may reflect changes in the generation and distribution of vulnerable
mass over the data carrier, not only changes in nearest-boundary scale
~\citep{zhang2019trades,wu2020_adversarial_weight_perturbation}.

\emph{Verification refinement is morphology-limited.}
Robustness verification is usually posed as a sound decision problem over a perturbation region, but
the finite-scale obstruction is not only whether vulnerable points are present.
At finite scale, the obstruction to robustness verification is whether its representation can separate unsafe geometry from safe geometry.
Theorem~\ref{thm:matched_scale_occupancy_law} isolates the corresponding local object in the present
theory: under matched rescaling,
\[
d_{\mathrm{BL}}\!\bigl(\omega_{r(t,h),t},\nu_h\bigr)\to0 .
\]
The limiting profile $\nu_h$ records how the newly vulnerable region is arranged on the normalized
template $K$.
Thus the relevant finite-scale object is a morphology of occupancy, not merely a scalar amount of
vulnerability.
A certificate may become difficult when this morphology is thin, interlaced, or mixed with other local
profiles in a way that cannot be separated by the current verification representation.

Different verification methods interact with this morphology at different levels.
Exact constraint-based methods, such as SMT or MILP formulations for piecewise-linear networks, can determine whether a specified perturbation region contains an unsafe input~\citep{katz2017reluplex,tjeng2019evaluating}. Repeated feasibility queries over refined regions could, in principle, provide increasingly localized information about the unsafe set.
Complete branch-and-bound verifiers refine this representation by splitting input regions or activation
regimes, and their refinement process can be understood as progressively exposing the relevant local
morphology~\citep{bunel2018unified}.
By contrast, abstract interpretation, convex relaxation, and linear bound propagation usually propagate
outer approximations of input neighborhoods, activation sets, or output bounds rather than the
occupancy profile itself
~\citep{wong2018provable,gehr2018ai2,singh2019deeppoly,zhang2022alphabeta_crown}.
The geometric obstruction is nevertheless common to all of these approaches.
If one verification element contains both safe and unsafe portions of the relevant local morphology, or
mixes several incompatible local regimes, then any sound coarse representation must retain all
possibilities until the element is refined.
The resulting conservatism is therefore not merely numerical; it reflects a failure to make the
finite-scale morphology of vulnerability visible.

\emph{Attack efficiency is delivery-limited.}
In iterative attacks, algorithmic progress is often monitored through surrogate loss decrease,
margin reduction, or the number of steps required to reach failure.
The realized theory separates this attack-side progress from progress relative to the reference
robustness geometry.
Theorem~\ref{thm:cell_delivery} identifies the local conversion rate from realized path length to
decrease of the ambient extension $\hat\phi$:
\[
\mathscr R_{\aaa\bbb}
=
\varrho^{k_\aaa^{\hat\phi}}\cos\Theta_{\aaa\bbb}.
\]
Thus descent under the observable attack objective is geometrically efficient only when it is delivered
as decrease of the ambient extension of the reference escape field.
Corollary~\ref{cor:reference_scale_realization_efficiency} gives the endpoint implication: if delivery
on the realization-side frontier segments is weak, or if non-frontier motion carries a large gauge
overhead, then successful escape may require a perturbation substantially larger than the reference
energy scale $\phi(x)$.
Attack efficiency depends on how effectively observable descent is converted into reference-side escape progress.

This gives a structural interpretation of gradient masking and unreliable attack evaluation.
A mechanism may fail to expose vulnerability at the relevant budget scale because the observable
geometry available to that attack is poorly aligned with the ambient escape geometry, rather than
because the vulnerability has been removed.
This is close in spirit to the failure mode isolated by obfuscated-gradient analyses
~\citep{athalye2018obfuscated}.
Reliable evaluation should therefore test more than one local observable regime and should be designed
to avoid common evaluation failures~\citep{carlini2019evaluating}.
Diverse attack suites and standardized evaluation protocols can be interpreted as attempts to probe
multiple delivery channels rather than relying on one surrogate geometry
~\citep{croce2020reliable,croce2021robustbench}.
Random restarts, momentum, adaptive step choices, stochastic proposal laws, and other history-dependent
modifications can be viewed as refinements of the observable mechanism.
They can be interpreted as attempts to move the realized path away from poorly delivered local
regimes or to stabilize trajectory segments with stronger conversion into reference-side progress, without changing the underlying escape geometry
as indicated by Remarks~\ref{rem:realization_side_refinements} and~\ref{rem:state_augmentation_mechanism}.

\emph{Transferability is persistence across readouts.}
In black-box transfer attacks, an adversarial example is optimized on a source or surrogate model and
then evaluated on an unseen target model.
High source-model attack strength does not by itself explain transfer.
A source trajectory may reduce the surrogate loss or margin efficiently, but transfer is structurally explained when the escape direction it has constructed remains effective under the target model's robustness geometry.
Remark~\ref{rem:persistence_of_delivered_decrease} formalizes this distinction.
In a nonconspiratorial reading of transfer, success is not attributed to an accidental agreement of
terminal failures across two models.
The structural object is instead the persistence of delivered decrease after the ambient reading
changes.
The regionwise analogue is shared attackability.
Nearby inputs may each be vulnerable while failing to support one common transferable mode; by
Theorem~\ref{thm:dominant_branch_leakage_law}, such coherence is lost when mass leaks from a dominant
observable branch into competing branches.
The alignment of a source-induced attack branch with the target geometry is therefore a central determinant of transferability.
This explains why the best white-box or source-side optimizer need not be the best transfer mechanism,
and why transfer performance is highly contextual across attacks, defenses, and target architectures
~\citep{zhao2026revisiting_transfer}.

The usual categories of transfer attacks can be interpreted as different strategies for preserving such
an attack branch under the source--target change.
Gradient stabilization regularizes the update history so that the perturbation does not overfit too
quickly to a source-specific local direction~\citep{dong2018boosting}.
Input augmentation exposes the attack trajectory to a family of transformed observations, thereby
favoring branches whose delivered effect persists across changes of the observable geometry rather
than branches tied to one source-specific local readout~\citep{xie2019improving}.
Feature disruption moves the attack objective from output-level class scores to intermediate
representations, where the induced geometry may be less target-specific.
Surrogate refinement modifies the source model so that the generated perturbation is shaped by a
geometry closer to likely targets.
Generative modeling learns a proposal field that can produce transferable perturbations without relying
solely on one local iterative trajectory.
From this perspective, transfer methods differ not only in empirical success rate, but in which part of
the attack pipeline they use to stabilize delivered escape across models.
This also clarifies why fair intra-category comparisons and transferability--iteration curves are
important: they test whether a method preserves a transferable branch, rather than merely increasing
source-model optimization strength~\citep{zhao2026revisiting_transfer}.

\emph{Mesoscopic nonclosure in large-capacity models.}
Macroscopic robustness behavior is mediated by mesoscopic organization.
Between pointwise margins and global robustness statistics lies a finite-resolution geometry that
determines how local escape structures aggregate into vulnerable mass, certification morphology,
realized delivery, and transferable branches.
Model complexity does not invalidate this layer; it makes its stability a substantive question.
The issue is whether one finite-resolution organization remains stable under refinements of scale or
changes of context, prompt, data region, or surrogate readout.
For highly expressive systems, this stability can become fragile.
Deep piecewise-linear networks can induce a rapidly growing number of linear regions, and expressive
deep architectures may organize the input space through many interacting local regimes
~\citep{montufar2014number,raghu2017expressive}.
Transformer architectures introduce self-attention and sequence-dependent representations, and their
expressive power has been analyzed theoretically
~\citep{vaswani2017attention,yun2020transformers}.
In the language of the present theory, increasing expressive capacity can fragment the mesoscopic
organization that mediates robustness, even though the local objects themselves remain well defined.

This phenomenon may be understood as an effective failure of mesoscopic closure: a fixed catalog of local generation types, occupancy
morphologies, delivery regimes, or branch histories may cease to be representative after small changes
of scale or readout.
Then robustness cannot be summarized by one dominant shell type, one occupancy morphology, one delivery
regime, or one attack branch.
Vulnerable mass, certification difficulty, realized attack efficiency, and transfer can vary sharply
because different active structures are exposed at neighboring resolutions or under neighboring
contexts.
Large language models provide a natural example of this caution, since adversarial suffix attacks show
that aligned models may possess mechanism-dependent and transferable failure channels
~\citep{zou2023universal}.
The general lesson is broader than that example: for large-capacity and context-dependent systems,
robustness analysis should test whether local generation laws, occupancy morphologies, delivery regimes,
and branch structures stabilize across scales and readouts before treating any single margin,
robustness-curve value, certificate, or attack configuration as representative.

\subsection{Relation to Geometric Views of Robustness}
\label{subsec:ml_relation_views}

Geometric and measure-theoretic accounts of robustness examine how vulnerability depends on
properties of the data distribution, model boundaries, adversarial directions, and learned
representations.
These approaches reveal complementary structures underlying robustness phenomena.
In the present framework, their connections can be expressed through the formation of escape
energy on a measured reference carrier, the growth and morphology of vulnerable mass, and the
delivery and persistence of realized escape.
The following discussion relates these established viewpoints to the corresponding components
of the theory.

\emph{Data-carrier and metric-measure geometry.}
Studies based on low-dimensional data manifolds examine the role of codimension and distinguish
adversarial perturbations confined to the data manifold from those realized in the ambient space
~\citep{khoury2019geometry,stutz2019disentangling,xiao2025understanding_onmanifold}.
Analyses of smooth generative models derive classifier-independent vulnerability bounds from
the geometry and dimension of the data-generating map
~\citep{fawzi2018adversarial_vulnerability}.
Metric-measure and concentration-based approaches likewise study vulnerability as a
phenomenon distributed across the data law
~\citep{gilmer2018adversarial_spheres,dohmatob2019generalized}.
Together, these results establish the importance of the geometric carrier and reference measure
in assessing robustness.

In the present theory, the pair $(\mathcal M,\mu)$ specifies the reference geometry and the
measure with respect to which vulnerability is evaluated.
The escape-energy field is organized on $\mathcal M$; shell bases and transverse openings
determine how newly vulnerable mass accumulates as the budget varies, while normalized
occupancy profiles describe its local spatial arrangement.
Admissible realization mechanisms specify the ambient paths along which that vulnerability
can be reached.
The distribution of vulnerable reference points and the geometry of realized attack paths
therefore form two connected levels of analysis.

\emph{Boundary and local score geometry.}
A second line of work studies the interaction between model scores, retention boundaries,
and the data distribution.
Linear explanations relate adversarial sensitivity to the accumulation of small
coordinate-wise changes in high dimension, while boundary-tilting analyses emphasize the
orientation and proximity of decision boundaries relative to the data geometry
~\citep{goodfellow2015explaining,tanay2016boundary}.
Studies of universal perturbations further connect vulnerability to boundary curvature and
correlated boundary structure across inputs
~\citep{moosavi2018robustness_universal}.
These approaches identify local score and boundary geometry as important determinants of
adversarial sensitivity.

In the escape formulation, this geometry enters through the escape-energy field and the
transverse opening of vulnerable sets near their active shell bases.
The componentwise growth law relates the local budget response of vulnerable mass to the
dimension of a shell base and the order of transverse rise.
Normalized occupancy profiles additionally record the finite-scale arrangement of newly
vulnerable regions.
Consequently, comparable pointwise margins or nearest-boundary distances may accompany
different mass-growth rates and occupancy morphologies.

\emph{Regional coherence and shared adversarial geometry.}
Universal perturbations and transferable adversarial subspaces reveal directions that remain
effective over substantial portions of the data distribution
~\citep{moosavi2017universal,tramer2017space_transferable}.
Theoretical analyses establish transferable adversarial directions for families of linear classifiers and certain two-layer ReLU networks
~\citep{charles2019geometric_transferability}.
These findings motivate a regional account of vulnerability in which common perturbation directions remain effective across multiple inputs or classifiers.

Within the present framework, regional coherence can have distinct geometric sources.
On the reference side, nearby points may share compatible generation types, transverse
openings, or shell organization.
On the realization side, an observable-guided attack branch may remain dominant across a
region of initial points.
The dominant-branch leakage law characterizes the loss of such shared realized behavior as
mass passes from a dominant branch into competing observable histories.
Shared adversarial directions can accordingly be studied through the organization of
possible escape and the persistence of delivered escape under a concrete mechanism.

\emph{Feature and representation geometry.}
A fourth line of work relates adversarial vulnerability to learned representations.
Accuracy--robustness tradeoff analyses and non-robust-feature accounts show how predictive
performance can rely on features that are informative under the data distribution yet
sensitive to admissible perturbations
~\citep{tsipras2019robustness,ilyas2019features}.
These accounts locate part of the explanation for vulnerability in the representation of
the retained task property.

The escape framework distinguishes several roles that such features may play.
Feature dependence may shape the reference escape-energy field by changing energy-neutral
directions or transverse rise, and may affect the finite-scale occupancy of newly vulnerable
regions.
Feature-level readouts may also define the observable geometry guiding realized attack
paths, while recurring feature directions across inputs or models may contribute to regional
coherence.
This places feature-based explanations within the geometry of escape generation,
morphology, delivery, and persistence.

\section{Concluding Remarks}
\label{sec:concluding_remarks}

A foundational measure-geometric theory of robustness as escape has been established here.
In this theory, a reference measure, a perturbation gauge, and an admissible realization mechanism jointly determine the structure of vulnerability.
The theory provides a common geometric language for developing more specific accounts of margins, certificates, attacks, transfer, and feature-based explanations.
Robustness is organized across energy, scale, reference geometry, observable mechanism, and realized path.
The central analysis jointly addresses the possibility of escape and the geometric processes through which vulnerability is formed, organized, delivered, and preserved.
A fundamental distinction lies between the escape possibilities encoded by the reference geometry and what an admissible mechanism can actually realize.
Robustness signatures are meaningful only insofar as the underlying geometry remains coherent.
When that geometry fragments, the signatures may still be observed, but their interpretive force must be re-examined.

% Acknowledgements and Disclosure of Funding should go at the end, before appendices and references

% \acks{The authors thank the reviewers, whose careful and insightful comments improve the
% presentation of the paper and clarify inconsistencies significantly.}


\vskip 0.2in
\bibliography{main}

%%%%%%%%%%%%%%%%%%%%%%%%%%%%%%%%%%%%%%%%%%%%%%%%%%%%%%%%

% \label{app:theorem}

% Note: in this sample, the section number is hard-coded in. Following
% proper LaTeX conventions, it should properly be coded as a reference:

%In this appendix we prove the following theorem from
%Section~\ref{sec:textree-generalization}:

% In this appendix we prove the following theorem from
% Section~6.2:

% \noindent
% {\bf Theorem} {\it Let $u,v,w$ be discrete variables such that $v, w$ do
% not co-occur with $u$ (i.e., $u\neq0\;\Rightarrow \;v=w=0$ in a given
% dataset $\dataset$). Let $N_{v0},N_{w0}$ be the number of data points for
% which $v=0, w=0$ respectively, and let $I_{uv},I_{uw}$ be the
% respective empirical mutual information values based on the sample
% $\dataset$. Then
% \[
% 	N_{v0} \;>\; N_{w0}\;\;\Rightarrow\;\;I_{uv} \;\leq\;I_{uw}
% \]
% with equality only if $u$ is identically 0.} \hfill\BlackBox

% \section{}

% \noindent
% {\bf Proof}. We use the notation:
% \[
% P_v(i) \;=\;\frac{N_v^i}{N},\;\;\;i \neq 0;\;\;\;
% P_{v0}\;\equiv\;P_v(0)\; = \;1 - \sum_{i\neq 0}P_v(i).
% \]
% These values represent the (empirical) probabilities of $v$
% taking value $i\neq 0$ and 0 respectively.  Entropies will be denoted
% by $H$. We aim to show that $\fracpartial{I_{uv}}{P_{v0}} < 0$....\\

% {\noindent \em Remainder omitted in this sample. See http://www.jmlr.org/papers/ for full paper.}

\end{document}